\documentclass[lettersize,journal]{IEEEtran}
\usepackage{tikz}
\usepackage{graphicx}
\usepackage{amsmath}
\usepackage{amssymb}
\usepackage{booktabs}
\usepackage{bm}
\usepackage{comment}
\usepackage{manyfoot}%
\usepackage{floatrow}
\usepackage{subfigure} 
\usepackage{multirow}
\usepackage[ruled,linesnumbered]{algorithm2e}
\usepackage{array}
\usepackage{stfloats}
\usepackage{textcomp}
\usepackage{url}
\usepackage{verbatim}
\usepackage{xcolor}
\usepackage{float}
\floatstyle{plaintop}
\restylefloat{table}
\usepackage[square]{natbib}
\setcitestyle{numbers,square}
\usepackage[pagebackref=true,breaklinks=true,letterpaper=true,colorlinks,bookmarks=false]{hyperref}
\usepackage[capitalize]{cleveref}

\hypersetup{
  colorlinks=true,breaklinks,
  linktoc=section,
  linkcolor=red,
  linkbordercolor=white,
  citecolor=blue,
  urlcolor=red,
}

\def\BibTeX{{\rm B\kern-.05em{\sc i\kern-.025em b}\kern-.08em
    T\kern-.1667em\lower.7ex\hbox{E}\kern-.125emX}}
\usepackage{balance}

\crefname{subsection}{Section}{Sections}
\Crefname{section}{Section}{Sections}
\crefname{table}{Table}{Tables}
\crefname{figure}{Figure}{Figures}
\crefname{equation}{Eqn.}{Eqns.}
\crefname{algorithm}{Algorithm}{Algorithms}

\def\R{\bm{\mathrm{R}}}

\def\degree{$^\circ$}
\def\etal{\textit{et al.}}
\def\eg{\textit{e.g.}}
\def\ie{\textit{i.e.}}

\def\imsc{0.9}
\def\fs{\small}

\def\tR{\bm{\tilde{\mathrm{R}}}}
\def\B{\bm{\mathrm{B}}}
\def\W{\bm{\mathrm{W}}}
\def\I{\bm{\mathrm{I}}}
\def\v{\mathfrak{r}}
\def\b{\bm{\mathrm{b}}}
\def\M{\bm{\mathrm{M}}}

\newcommand{\PreserveBackslash}[1]{\let\temp=\\#1\let\\=\temp}
\newcolumntype{C}[1]{>{\PreserveBackslash\centering}p{#1}}
\newcolumntype{R}[1]{>{\PreserveBackslash\raggedleft}p{#1}}
\newcolumntype{L}[1]{>{\PreserveBackslash\raggedright}p{#1}}

\begin{document}
\title{EAR-Net: Pursuing End-to-End Absolute Rotations from Multi-View Images}

\author{Yuzhen Liu, Qiulei Dong
\thanks{
Yuzhen Liu is with the State Key Laboratory of Multimodel Artificial Intelligence Systems, 
Institute of Automation, Chinese Academy of Sciences, Beijing 100190, China, and 
also with the School of Artificial Intelligence, University of Chinese Academy of Sciences, Beijing 10049, China
(e-mail: liuyuzhen2022@ia.ac.cn)

Qiulei Dong is with the State Key Laboratory of Multimodel Artificial Intelligence Systems, 
Institute of Automation, Chinese Academy of Sciences, Beijing 100190, China, 
also with the School of Artificial Intelligence, 
University of Chinese Academy of Sciences, Beijing 10049, China, 
and the Center for Excellence in Brain Science and Intelligence Technology, 
Chinese Academy of Sciences, Beijing 100190, China
(e-mail: qldong@nlpr.ia.ac.cn). (Corresponding author: Qiulei Dong.)
}
}

\markboth{}%
{How to Use the IEEEtran \LaTeX \ Templates}

\maketitle

\begin{abstract}
Absolute rotation estimation is an important topic in 3D computer vision. Existing works in literature generally employ a multi-stage (at least two-stage) estimation strategy where multiple independent operations (feature matching, two-view rotation estimation, and rotation averaging) are implemented sequentially. However, such a multi-stage strategy inevitably leads to the accumulation of the errors caused by each involved operation, and degrades its final estimation on global rotations accordingly. To address this problem, we propose an End-to-end method for estimating Absolution Rotations from multi-view images based on deep neural Networks, called EAR-Net. The proposed EAR-Net consists of an epipolar confidence graph construction module and a confidence-aware rotation averaging module. The epipolar confidence graph construction module is explored to simultaneously predict pairwise relative rotations among the input images and their corresponding confidences, resulting in a weighted graph (called epipolar confidence graph). Based on this graph, the confidence-aware rotation averaging module, which is differentiable, is explored to predict the absolute rotations. Thanks to the introduced confidences of the relative rotations, the proposed EAR-Net could effectively handle outlier cases. Experimental results on three public datasets demonstrate that EAR-Net outperforms the state-of-the-art methods by a large margin in terms of both accuracy and inference speed. 
\end{abstract}

\begin{IEEEkeywords}
Camera Pose Estimation, Rotation Averaging, Deep Learning
\end{IEEEkeywords}

\section{Introduction}
\label{sec:intro}

Absolute rotation estimation is a challenging and important topic in 
computer vision, which is to calculate the absolute or global camera 
rotations from a set of multi-view images. It has a broad range of applications, 
including structure from motion (SfM) \cite{colmap,dong2022}, 
multiple view stereo (MVS) \cite{mvsnet,bae2022multi}, 
novel view synthesis \cite{mildenhall2021nerf},
\textit{etc.}

\begin{figure*}[t]
\centering
\includegraphics[width=\linewidth,clip]{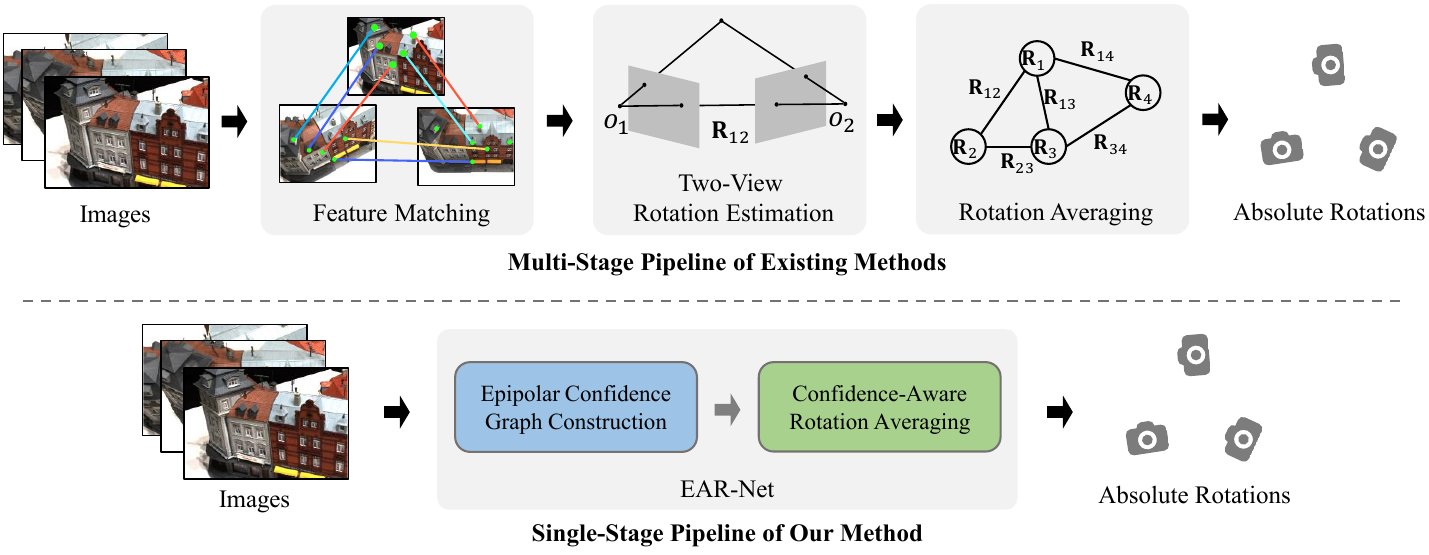}
\caption{
Comparison of absolute rotation estimation pipelines between the existing methods and the proposed  EAR-Net.
\textbf{Top:} Multi-stage pipeline of the existing methods in literature, which performs  
feature matching, two-view rotation estimation, and rotation averaging respectively at multiple sequential stages.
\textbf{Bottom:} Single-stage pipeline of the proposed EAR-Net 
(consisting of an epipolar confidence graph construction module and a confidence-aware rotation averaging module)
which estimates the absolute rotations in an end-to-end manner. 
}
\label{fig:traditional}
\end{figure*}

As shown in \Cref{fig:traditional} (top), a widely-used strategy for 
absolute rotation estimation in literature includes the 
following multiple sequential but independent stages 
(at least two independent stages, \ie, the following stages (a) 
and (b) are combined into one stage `end-to-end relative rotation estimation' in 
\cite{melekhov2017relative,en2018rpnet,chen2021wide,Cai2021extreme}): 
(a) Feature matching: A traditional manner in literature is firstly to 
implement a feature detector \cite{SIFT,keynet} for 
extracting keypoints from each input image, and then build descriptors 
\cite{R2D2,HyNet} for these  keypoints, and finally match 
these points among the input images. 
Recently, a popular manner is to design a feature matching network 
for directly outputting the point correspondences without an 
explicit detector nor an explicit descriptor 
\cite{LoFTR,aspanformer}.  
(b) Two-view rotation estimation: Once the feature correspondences are 
obtained among the input images, the classic 5-point algorithm 
\cite{nister2004efficient} (jointly with RANSAC \cite{RANSAC} in 
most cases for alleviating the influences of outliers)  
is implemented to calculate the relative rotation between each 
pair of images.
(c) Rotation averaging: Once all the relative rotations 
(\ie, two-view rotations) are obtained, the absolute 
rotations corresponding to all the cameras are calculated by 
implementing a rotation averaging algorithm 
\cite{HARA,DISCO,IRLS,Chen_2021_CVPR}.   

However, the aforementioned multi-stage strategy inevitably leads 
to the accumulation of error generated at each stage, which is a 
common problem confronted by many multi-stage methods for handling 
various visual tasks \cite{chen2020graph,IRLS,detr}. 
Moreover, even if the performance at one stage 
is improved, it could not guarantee an improved performance at its subsequent stage. 
For example, a larger number of correct matches does not 
necessarily lead to a better estimation of relative rotations, 
as indicated by Fan \etal{} \cite{RANSACRole}.
We have further conducted an experiment to verify this issue, 
and please see more details in the Appendix.

Inspired by the success of end-to-end learning-based strategies 
in many other visual tasks \cite{dsac,epro} for avoiding 
the above problems that multi-stage methods have to be confronted with, 
the following problem is naturally raised: 
\textit{Is it possible to combine the aforementioned stages into only one 
stage and estimate the absolute rotations from multi-view 
images in an end-to-end manner for pursuing a better performance?}
It is noted that it is not trivial to deal with this problem, 
since the process of identifying and filtering outliers generated 
at the aforementioned Stage (b) (feature match outliers) and Stage (c) 
(relative rotation outliers) is non-differentiable and 
hinders end-to-end learning. 
Here, two additional points have to be explained: 
(1) It is theoretically feasible to simply combine an end-to-end relative 
rotation estimation sub-network and an end-to-end rotation averaging sub-network 
into an end-to-end network for computing absolute rotations from multi-view images, 
however, such a simple combination is less competitive as 
demonstrated by the experimental results in \cref{subsec:concat}. 
(2) The incremental SfM technique \cite{snavely2006photo,colmap} 
could recover the rotations by joint optimization of poses and scene 
depths together from multi-view images, however, it does not mean 
incremental SfM is opposed to the absolute rotation estimation technique. 
In fact, compared with incremental SfM that generally has to 
spend relatively expensive costs on implementing BA (bundle adjustment) multiple times, 
the absolute rotation estimation technique generally spends less computational cost.
Moreover, the estimated rotations could be 
applied to incremental SfM approaches for reducing drift \cite{Chen_2021_CVPR,allinweights}.
Hence, we do not give a further analysis and 
comparison between the two techniques in the following parts.

To address the above issues, we propose an end-to-end method for 
absolute rotation estimation from multi-view images in this paper, 
called \textbf{EAR-Net}. 
As shown in \cref{fig:traditional} (bottom), 
the proposed EAR-Net consists of an epipolar confidence graph construction (ECGC)
module and a confidence-aware rotation averaging (CARA) module.
The ECGC module is explored to use the input multi-view images to 
learn pairwise relative camera rotations and their confidences. 
Accordingly, a weighted graph (called epipolar confidence graph) is built, 
whose vertices are the absolute rotations and edges are the pairwise relative 
rotations weighted by the learned confidences. 
A relative rotation with a low confidence is considered to be less 
accurate than that with a high confidence.
Then, the CARA module is explored to use the built epipolar confidence graph for 
learning the absolute rotations. 
In the explored CARA module, a confidence-aware loss is designed for assigning different 
weight to different relative rotations, and simultaneously alleviating the 
negative influence of outliers, by utilizing the confidence scores from the built 
epipolar confidence graph in the ECGC module. 
And an iterative optimization algorithm, which is differentiable, is presented 
to minimize the confidence-aware loss, so that the proposed EAR-Net could be 
trained in an end-to-end manner.

To summarize, our main contributions include:
\begin{itemize}
\item We explore the ECGC module for building an epipolar confidence graph from an 
input set of multi-view images. The ECGC module could not only learn pairwise relative 
camera rotations, but also their degrees of confidence for reflecting whether each 
estimated relative rotation is accurate enough.

\item We explore the CARA module with a designed confidence-aware loss for 
predicting the absolute camera rotations from the built epipolar confidence graph. 
And an iterative optimization algorithm is introduced for 
optimizing the confidence-aware loss accordingly.

\item By integrating the aforementioned ECGC and CARA modules together, 
we proposed the EAR-Net for absolute rotation estimation. 
Thanks to the introduced optimization algorithm in the second contribution, 
the proposed EAR-Net could be trained in an end-to-end manner. 
To the authors' best knowledge, this work is the first attempt to unify relative 
rotation estimation and rotation averaging in an end-to-end manner, 
and its superiority to some state-of-the-art multi-stage methods has 
been demonstrated by the experimental results in \cref{sec:experiments}. 
\end{itemize}

This paper is organized as follows: 
\cref{sec:related_work} gives a review of existing relative rotation estimation 
and rotation averaging methods.
\cref{sec:method} introduces the details of the proposed EAR-Net.
In \cref{sec:experiments}, we conduct experiments on public datasets 
to demonstrate the effectiveness of EAR-Net.
\cref{sec:conclusion} concludes the paper.

\section{Related Work}
\label{sec:related_work}

In this section, we review the existing works on relative rotation 
estimation, rotation averaging, and absolute rotation estimation respectively.

\subsection{Relative Rotation Estimation}

Relative rotation estimation is to calculate the camera rotation 
between an input pair of images. A traditional strategy for 
relative rotation estimation contains the following two key stages: 
feature matching 
\cite{SIFT,R2D2,LoFTR,matchformer} 
and two-view rotation estimation via the 5-point 
algorithm \cite{nister2004efficient}. 
However, it is hard for such a strategy to obtain a reliable estimation 
on relative rotation in the cases of weak textures or degenerated 
configurations \cite{RANSACRole}. 

Unlike the above strategies, several recent works have adopted deep 
neural networks (DNN) to estimate relative 
rotations in an end-to-end manner
\cite{melekhov2017relative,en2018rpnet,chen2021wide,Cai2021extreme}. 
For example, Siamese architectures were employed for relative pose 
estimation in \cite{melekhov2017relative,en2018rpnet}. 
Zhou \etal{} \cite{rot6d} gave an analysis on the continuity of rotation 
representations, and suggested representing 3D rotations in 
a 6D space, which benefits the training of neural networks. 
Different from the above regression-based 
methods \cite{melekhov2017relative,en2018rpnet}, 
Cai \etal{} \cite{Cai2021extreme} cast the relative rotation 
estimation problem as a classification problem, where each class 
represents an angle in the range [-180$^\circ$, 180$^\circ$]. 
However, it is worth noting that the above DNN-based methods are 
only available for estimating the relative rotation from an input 
pair of images, but could not estimate the absolute rotations 
from multi-view images, which are significantly different from our method.

\subsection{Rotation Averaging}

Rotation averaging is a widely studied task in computer vision, 
which aims to recover the absolute rotations of cameras 
from a given set of relative rotations. 
It plays an important role in global SfM methods 
\cite{hartleyAvg,IRLS,cui2018voting,Chen_2021_CVPR},
which is efficient for speeding up camera pose estimation, 
and could also reduce camera drift \cite{Chen_2021_CVPR}.
The early method by Govindu \etal{} \cite{govindu2004lie} 
used a Lie-algebra representation to average relative rotations in a 
least-squares manner. To improve 
robustness, several works have either focused on using robust 
loss functions \cite{hartleyAvg,IRLS} or removing outliers before 
the optimization \cite{govindu2006robustness,zach2010disambiguating,HARA}. 
For example, Hartley \etal{} \cite{hartleyAvg} proposed an $\ell_1$ 
averaging method based on the Weiszfeld algorithm, considering that 
the $\ell_1$ norm was more robust to outliers than the $\ell_2$ norm. 
Chatterjee \etal{} \cite{IRLS} proposed a generalized framework where 
different robust loss functions could be seamlessly embedded, and the 
optimization could be implemented in an efficient iteratively re-weighted 
least-squares (IRLS) manner in the Lie-algebra representation. 
Govindu \etal{} \cite{govindu2006robustness} used a RANSAC-based approach 
to detect and remove erroneous relative rotations. 
Zach \etal{} \cite{zach2010disambiguating} proposed to filter 
outlier edges by using loop constraint, \textit{i.e.}, 
chaining all noise-free transformations 
along a loop should result in an identity transformation. 
Lee \etal{} \cite{HARA} proposed an initialization scheme based on a hierarchy of
triplet support, and removed edges that do not conform to an 
initial solution. Moreover, some works found that a good initialization 
approach is also helpful for improving the rotation averaging accuracy 
\cite{IRLS,cui2018voting,Chen_2021_CVPR,gao2021incremental,HARA}. 
Sidhartha \etal{} \cite{allinweights} extensively analyzed the performance of 
IRLS-based rotation averaging \cite{IRLS,Chen_2021_CVPR,HARA}, 
and further pointed out that the performance of 
these methods are intimately related to both the robust 
loss function and initialization approach of the absolute rotations.

Additionally, some learning-based rotation averaging methods 
have been proposed by utilizing deep networks 
recently \cite{NeuRoRA,RAMSP,RAGO}. Here, it has to be explained that these 
learning-based rotation averaging methods take relative 
rotations as inputs. However, different from the above methods, the proposed 
EAR-Net recovers the absolute rotations directly from the original images. 

\subsection{Absolute Rotation Estimation}

Absolute rotation estimation \cite{wang2023posediffusion,lin2024relpose++,taiana2024prago} 
is to recover the camera absolute rotations from a set of images.
For example, PoseDiffusion \cite{wang2023posediffusion} 
introduces a diffusion-based strategy to recover the absolute camera poses from a set of images. 
RelPose++ \cite{lin2024relpose++} first estimates a distribution over relative poses and 
then lifts them to absolute poses. 
PRAGO \cite{taiana2024prago} is a closely related differentiable method for 
absolute rotation estimation. It employs a two-stage strategy: 
Firstly, it uses the 5-point algorithm to estimate the pairwise relative 
poses according to the input set of images with matching objectness detections, 
and this stage is non-differentiable. 
Secondly, it refines the obtained pairwise relative poses and estimates the 
rotations by jointly optimizing a refinement loss and an averaging loss in a differentiable manner. 
Similar to the other multi-stage methods, PRAGO has to suffer from the error 
accumulation problem to some extent. In addition, it has to be pointed out that PRAGO 
takes a set of images with matched objectness detections as its input, 
this is to say, it has to employ additional methods \cite{SuperGlue,kim2021oln} 
to obtain these matched objectness detections in advance. 

In addition, several pre-trained large foundation models \cite{wang2025continuous,wang2024dust3r,leroy2024mast3r,wang2025vggt}, 
which are generally trained on various datasets, have been proposed for simultaneously handling multiple visual tasks (including the absolute rotation estimation task) recently. 
However, since they have a large number of parameters, their inference time 
is generally much longer than their task-specific counterparts as 
verified by the results in \cref{subsec:evaluation}.

\begin{figure*}[t]
	\centering
	\begin{tikzpicture}
	\node[anchor=south west,inner sep=0] (image) at (0,0)
	{\includegraphics[width=0.95\linewidth,clip]{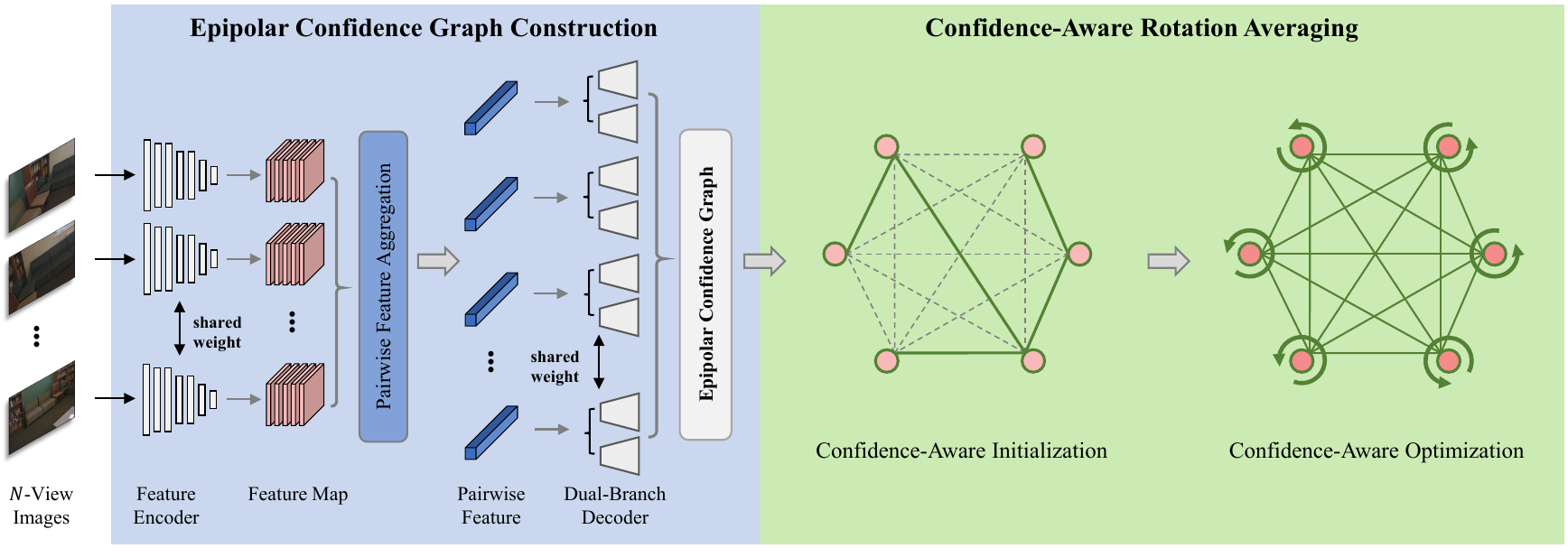}};
	\node[anchor=center,scale=0.7] at (15., 0.65) (wls) {(\Cref{alg:wls})};  
	\end{tikzpicture}
	\caption{Architecture of the proposed EAR-Net, which consists of the epipolar confidence 
	graph construction module for learning relative rotations and their confidences, 
	and the confidence-aware rotation averaging module for predicting the absolute 
	rotations. 
 }
	\label{fig:architecture}
\end{figure*}

\section{Methodology}
\label{sec:method}

In this section, we introduce the proposed EAR-Net for absolute rotation
estimation in detail.
Firstly, we introduce the architecture of the proposed EAR-Net.
Then, we introduce the key modules and the training strategy.
Finally, we introduce the strategy for extending our method to large-scale scenes.

\subsection{Architecture}\label{subsec:EAR-Net}

As shown in \Cref{fig:architecture}, the proposed EAR-Net takes a set of
multi-view images as inputs, and it aims to output the corresponding
absolute rotations. EAR-Net consists of the
epipolar confidence graph construction module and
the confidence-aware rotation averaging module.
In the proposed EAR-Net, the referred operations (including relative
rotation estimation, rotation averaging, \textit{etc.}) for estimating
absolute camera rotations are combined into one stage, and relative
camera rotations are only considered as intermediate features, but not
as final prediction results. It is noted that no matter an end-to-end
manner or a multi-stage manner is employed to learn relative rotations,
the learned relative rotations from different pairs of images generally
have different noise levels. Hence, given an input set of images, the
epipolar confidence graph construction module is designed to
learn the relative rotations and their confidences that indicate whether
the corresponding relative rotations are accurate enough, and then
a weighted epipolar confidence graph is constructed
according to the obtained relative rotations and confidences.
Based on the constructed epipolar confidence graph,
the confidence-aware rotation averaging module is designed to estimate
the absolute rotations. We will introduce the two modules,
the loss function and training strategy
in detail in the following subsections.

\subsection{Epipolar Confidence Graph Construction Module}\label{subsec:graph}

Given an arbitrary set of $N$-view images $\{\mathcal{I}_k\}_{k=1}^N$ with the height $H$ and width $W$,
\ie{}, $\mathcal{I}_k\in\mathbb{R}^{3\times H\times W}$,
the epipolar confidence graph construction module is designed to
simultaneously learn the relative camera
rotations and their corresponding confidences, respectively.
It consists of a feature encoder, a pairwise feature aggregation unit, and 
a dual-branch decoder:

\noindent\textbf{Feature Encoder. }
The feature encoder is to learn feature maps from the input $N$ images.
In this work, we use the first three residual blocks from ResNet18 
\cite{ResNet} as the feature encoder,
and it outputs feature maps $\{\mathcal{F}_k\}_{k=1}^N$ of 256 channels with
the size of $H/16\times W/16$, \ie, $\mathcal{F}_k\in \mathbb{R}^{256 \times H/16\times W/16}$.

\begin{figure*}
    \centering
    \includegraphics[width=0.9\linewidth]{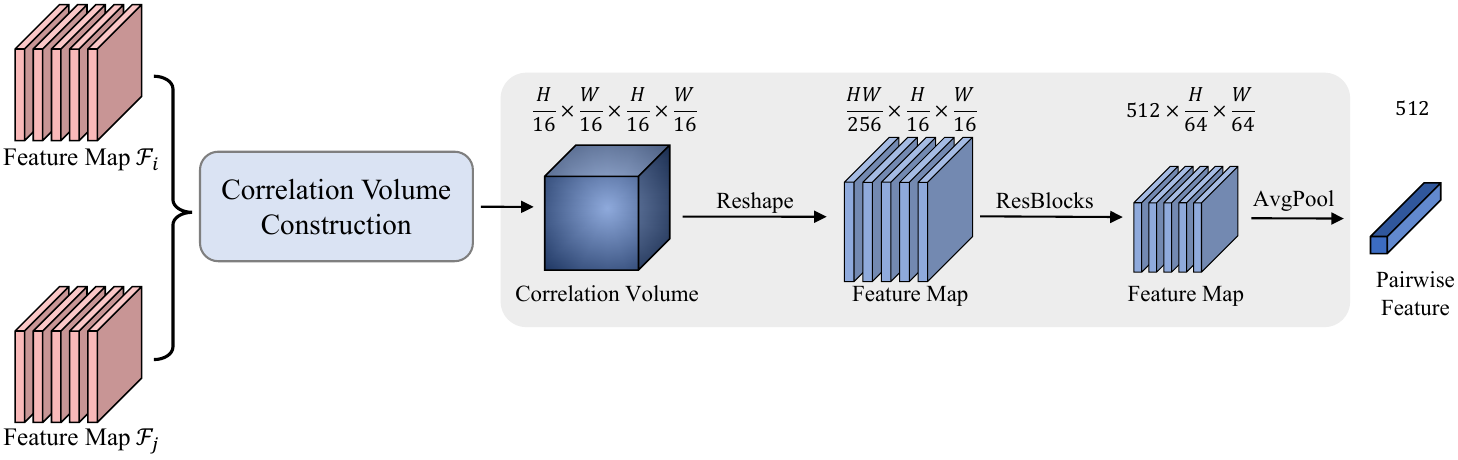}
    \caption{Architecture of the pairwise feature aggregation (PFA) unit. 
    This unit takes two image feature maps as input and outputs the 
    pairwise feature vector.}
    \label{fig:pfa}
\end{figure*}

\noindent\textbf{Pairwise Feature Aggregation. }
Once the feature maps for all the involved $N$ images are obtained by the feature encoder, 
the Pairwise Feature Aggregation (PFA) unit is explored to aggregate them pairwisely.
The architecture of the PFA unit is illustrated in \cref{fig:pfa}.
As seen from this figure, the PFA unit takes a pair of feature maps as 
inputs, and it outputs an aggregated feature vector.
Specifically, we first compute the 4D correlation volume 
$\bm{\mathrm{Q}}_{ij}\in\mathbb{R}^{H/16\times W/16\times H/16\times W/16}$
to encode the mutual information between the input pair of feature maps 
$\mathcal{F}_i$ and $\mathcal{F}_j$ as:
\begin{equation}\label{eqn:cv}
  \bm{\mathrm{Q}}_{ij}(h_1,w_1,h_2,w_2)=
  \sum\limits_{k=1}^{256}\mathcal{F}_i(k,h_1,w_1)\mathcal{F}_j(k,h_2,w_2)
\end{equation}
\noindent where $k,h_1,w_1,h_2,w_2$ are the dimension indices.

Then, the obtained 4D correlation volumes are reshaped as 
3D feature maps with size $\mathbb{R}^{HW/256\times  H/16\times W/16}$, 
and are passed into a residual block and an average pooling layer to 
obtain a feature vector (namely the pairwise feature) with the size of 512.

\begin{figure*}
    \centering
    \includegraphics[width=0.76\linewidth]{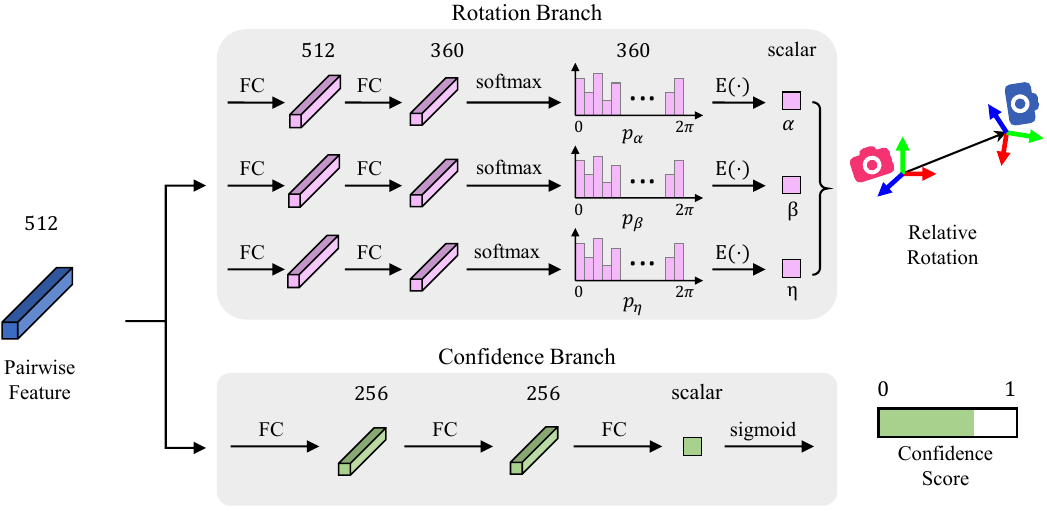}
    \caption{Architecture of the dual-branch decoder, which consists of 
    the rotation branch to predict the relative rotation, and 
    the confidence branch to predict the corresponding confidence.
    `FC' denotes the fully connected layer.
    $\alpha,\beta,\gamma$ denote the roll, pitch, and yaw of the three 
    Euler angles respectively, and 
    $p_\alpha,p_\beta,p_\eta$ denotes their probability distributions. 
    }
    \label{fig:decoder}
\end{figure*}

\noindent\textbf{Dual-Branch Decoder. }
The dual-branch decoder takes the pairwise features as
inputs, and it outputs the relative rotations and the corresponding
confidences. As shown in \Cref{fig:decoder}, the explored
dual-branch decoder consists of two branches: a
\textit{rotation branch} and a \textit{confidence branch}. 
The pairwise feature vectors are inputted to the two branches respectively.

\textit{Rotation branch:}
This branch maps each pairwise feature into
a relative rotation, which is parameterized with Euler angles
$\alpha$, $\beta$, $\eta$ (roll, pitch, yaw). Similar to
\cite{Cai2021extreme}, the distribution of $\alpha$, $\beta$, $\eta$ are
represented by $B$-dimensional vector $p_\alpha$, $p_\beta$, $p_\eta$ respectively 
(we set $B=360$),
where each element in the vectors indicates a probability
corresponding to an angle in [$0$, $2\pi$]. However, it
is non-differentiable to directly choose the angle with the maximum
probability as the prediction result as done in \cite{Cai2021extreme}.
Instead, we compute the following expectation as the final prediction:
\begin{equation}\label{eqn:expectation}
  \mathrm{E}(\theta)=\sum\limits_{k=1}^{B}p(\theta_k)\theta_k
\end{equation}
\noindent where $\theta_k$ denotes the angle represented by
the $k$-th element and $p(\theta_k)$ denotes the corresponding probability.

Specifically, as shown in \Cref{fig:decoder}, the rotation
branch firstly adopts two fully connected layers and a \texttt{softmax} function
to obtain the discrete distribution $p_\alpha$, $p_\beta$, $p_\eta$ for 
roll, pitch, and yaw respectively.
Then, the values of roll, pitch, and yaw are computed according to 
the expectation operation in \Cref{eqn:expectation}.  
Finally, given the three Euler angles, the corresponding
rotation matrix could be straightforwardly obtained.

\textit{Confidence branch:}
This branch consists of three fully connected layers and a 
\texttt{sigmoid} activation function. It maps each pairwise 
feature vector into a scalar (confidence) in the range of
[0, 1], which is expected to reflect the degree of confidence
on whether the estimated relative rotation in the above rotation
branch is accurate enough. This is to say, when the
confidence branch is trained, it tends to predict
large confidence scores for reliable relative rotations, and 
small confidence scores for unreliable relative rotations.

Once both relative rotations $\R_{ij}$ and their corresponding
confidences $c_{ij}$ are obtained through the above two branches,
the \textit{epipolar confidence graph} $\mathcal{G}=\{\mathcal{V},\mathcal{E}\}$
which is a weighted graph is constructed as follows:
Vertex $i\in\mathcal{V}$ represents the $i$-th camera with
an unknown absolute rotation $\R_i$;
Edge $(i,j)\in\mathcal{E}$ represents the relative rotation
$\R_{ij}$ between the $i$-th and $j$-th images,
which is weighted by the confidence $c_{ij}$.

\subsection{Confidence-Aware Rotation Averaging Module}\label{subsec:optimization}

Given the constructed epipolar confidence graph $\mathcal{G}=\{\mathcal{V},\mathcal{E}\}$
from the above module, the confidence-aware rotation averaging module aims to recover
the absolute rotations in an iterative and differentiable manner, which allows the 
gradient to back-propagate.

\noindent\textbf{Confidence-Aware Loss. }
Considering that the loss function is one of the main influencing factors for
rotation averaging \cite{allinweights},
we propose the confidence-aware loss (CAL) that is based on the learned confidence:

\begin{equation}\label{eqn:objective}
  \mathcal{L}_\mathrm{CAL}=\sum\limits_{(i,j)\in\mathcal{E}}c_{ij}\mathfrak{R}^2(\R_{ij},\R_j\R_i^\mathrm{T})
\end{equation}

\noindent where `$\mathfrak{R}(\cdot,\cdot)$' is the Riemannian distance:
$\mathfrak{R}(\bm{\mathrm{X}},\bm{\mathrm{Y}})=\|\log(\bm{\mathrm{X}}\bm{\mathrm{Y}}^\mathrm{T})\|_2$,
`$\|\cdot\|_2$' denotes the $\ell_2$ norm and `$\mathrm{log}(\cdot)$' denotes the mapping
from the Lie group to its Lie algebra.

Many hand-crafted loss functions 
(such as the Cauchy loss, Geman-McClure loss, \textit{etc.}) 
are adopted in other methods \cite{IRLS,HARA}, which are manually designed  
according to noise distribution in data \cite{NeuRoRA}.
In contrast, our approach explores direct learning of the loss function from data.
It should be pointed out that these hand-crafted robust loss functions 
could also be used seamlessly in our framework. However, 
we will show in \cref{subsec:ablation} that
our learning-based CAL gives a significantly better result.

To minimize the confidence-aware loss, we propose the \textit{Confidence-Aware Initialization} 
(CAI) approach and the \textit{Confidence-Aware Optimization} (CAO) algorithm, which 
will be introduced in the following.

\noindent\textbf{Confidence-Aware Initialization. }
Considering that an effective set of initial absolute rotations is generally important
for pursuing a more accurate final prediction, unlike the existing
initialization methods \cite{Chen_2021_CVPR,gao2021incremental,shen2016graph,HARA,allinweights}
that calculate the initials of absolute rotations according to
manually defined criteria (e.g. number of inlier matches
\cite{Chen_2021_CVPR,gao2021incremental,HARA} or similarity scores
\cite{shen2016graph}), the proposed CAI approach is based on the automatically learned
confidence. Specifically, we first construct a maximum spanning tree via the
classic Prim's algorithm according to the learned confidence.
Then, the initials $\{\R_i\}_{i=1}^N$  are computed by multiplying
the predicted relative rotations from the root vertex progressively
based on the maximum spanning tree.

\noindent\textbf{Confidence-Aware Optimization. }
Once the absolute rotations are initialized, we optimize them in an iterative manner.
As done in \cite{govindu2004lie}, in each iteration, we use the 
Euclidean distance in Lie algebra to
approximate the Riemannian distance in its Lie group, \ie,
$\mathfrak{R}(\bm{\mathrm{X}},\bm{\mathrm{Y}})\approx\|\log(\bm{\mathrm{X}})-\log(\bm{\mathrm{Y}})\|_2$.
Different from \cite{govindu2004lie} that solves the least squares problem corresponding to 
the $\ell_2$ loss in each iteration,
our method solves a weighted least squares problem corresponding to the introduced CAL.
In the following, we will introduce the algorithm (outlined in \Cref{alg:wls}) in detail.

Let $\I\in\mathbb{R}^{3\times 3}$ be the identity matrix,
$\{\tR_{i}\}_{i\in\mathcal{V}}$ be the set of current estimate of absolute rotations,
$\{\R_{ij}\}_{(i,j)\in\mathcal{E}}$ be the set of predicted relative rotations from the ECGC module,
$\{\Delta \v_i\}_{i\in\mathcal{V}}$ be the set of update amount we need to estimate,
$\{\Delta \b_{ij}\}_{(i,j)\in\mathcal{E}}$ be the set of residuals computed as 
$\Delta \b_{ij}=\log(\tR_j^\mathrm{T}\R_{ij}\tR_i)$.
Then, for edge $(i,j)$, we have:

\begin{equation}\label{eqn:B}
  \underbrace{\left[\cdots,\I,\cdots,-\I,\cdots\right]}_{\B_{ij}}\Delta \v = \Delta \b_{ij}
\end{equation}

\noindent where $\B_{ij}\in\mathbb{R}^{3\times 3N}$ is a block 
matrix where the $i$-th block is $-\I$,
the $j$-th block is $\I$, and all other elements are zero.

Stacking all $\B_{ij}$, $\Delta \b_{ij}$, $\Delta \v_i$ along the column, we obtain
$\B\in\mathbb{R}^{\frac{3N(N-1)}{2}\times 3N}$,
$\Delta \b\in\mathbb{R}^{\frac{3N(N-1)}{2}\times 1}$,
$\Delta \v\in\mathbb{R}^{3N\times 1}$ respectively.
Then minimizing the CAL in \Cref{eqn:objective} is equivalent to solving the following problem:
\begin{equation}\label{eqn:wls}
  \min_{\Delta\v} (\B\Delta\v-\Delta \b)^\mathrm{T}\W(\B\Delta\v-\Delta \b)
\end{equation}
where $\W$ is a diagonal matrix constructed as follows: 

\begin{equation}\label{eqn:W}
  \W=\left[\begin{array}{cccc}
      c_{12}\I &          &        &             \\
               & c_{13}\I &        &             \\
               &          & \ddots &             \\
               &          &        & c_{N-1,N}\I \\
    \end{array}\right]
\end{equation}

The solution to \cref{eqn:wls} is $\Delta\v=(\B^\mathrm{T}\W\B)^{-1}\B^\mathrm{T}\W\Delta\b$.
Then the absolute rotations are updated as 
$\tR_i\leftarrow \tR_i\exp{(\Delta \v_i)},\forall i\in\mathcal{V}$, where
`$\exp(\cdot)$' is the mapping from Lie algebra to its Lie group.

\begin{algorithm}[t]
	\small
	\caption{Confidence-Aware Optimization (CAO)}
	\label{alg:wls}
	\KwIn{
	Relative rotations $\{\R_{ij}\}_{(i,j)\in\mathcal{E}}$; 
	Initial absolute rotations $\{\tR_i\}_{i\in\mathcal{V}}$;
	Confidence $\{c_{ij}\}_{(i,j)\in\mathcal{E}}$; 
	Maximum number of iterations $T$}
	\KwOut{Estimated absolute rotations $\{\R_i\}_{i\in\mathcal{V}}$}  
	\BlankLine
	
	
	Construct the matrix $\B$ according to \Cref{eqn:B}
	
	Construct the block diagonal matrix $\W$ according to \Cref{eqn:W}

	$\M\leftarrow (\B^\mathrm{T}\W\B)^{-1}\B^\mathrm{T}\W,k\leftarrow 0$ 

	\While{$k<T$}{
		$\Delta \R_{ij}\leftarrow \tR_j^\mathrm{T}\R_{ij}\tR_i$

		$\Delta \b_{ij}\leftarrow\log(\Delta \R_{ij})$

		Gather $\Delta \b_{ij}$ into the residual vector $\Delta \b$

		$\Delta \v\leftarrow\M \Delta\b$

		$\tR_i\leftarrow\tR_i\exp(\Delta \v_i)$,  $\forall i\in\mathcal{V}$

		$k\leftarrow k+1$
	}
	
	$\R_i\leftarrow\tR_i,\forall i\in\mathcal{V}$
\end{algorithm}

\subsection{Training Strategy and Loss Function}\label{subsec:loss}

Here, we introduce the training strategy and loss function
used in the proposed EAR-Net as follows:

\noindent\textbf{Pretraining. }
We firstly pretrain the feature encoder, the PFA unit and the rotation 
branch of the dual-branch decoder together,  using 
ground-truth relative rotations as supervision signals, while the confidence 
branch is omitted. As indicated in \Cref{subsec:graph},
the Euler angle parameterization is used in the rotation 
branch. Hence, the loss function $\mathcal{L}_\mathrm{P}$ at this
pretraining stage consists of three cross-entropy
loss terms for roll ($\alpha$), pitch ($\beta$) and yaw ($\eta$)
respectively:

\begin{equation}\label{eqn:loss_stage1}
  \mathcal{L}_\mathrm{P} = \mathcal{L}_\mathrm{CE}(p_\alpha,p_{\hat{\alpha}})+
  \mathcal{L}_\mathrm{CE}(p_\beta,p_{\hat{\beta}})+
  \mathcal{L}_\mathrm{CE}(p_\eta,p_{\hat{\eta}})
\end{equation}
\noindent where $\mathcal{L}_\mathrm{CE}(\cdot,\cdot)$ is the cross
entropy loss, and $p_{\hat{\alpha}}$, $p_{\hat{\beta}}$, $p_{\hat{\eta}}$
denote the ground-truth distribution (one-hot vector) for roll, pitch and yaw respectively.

\noindent\textbf{End-to-End Training. }
After the feature encoder and the rotation branch of the dual-branch decoder are
initialized with the pretraining weights, 
the whole model of EAR-Net is trained in an end-to-end manner.
To remove the gauge ambiguity, we first obtain the transformation
between the estimated rotations $\{\R_{i}\}_{i=1}^N$ to the ground-truth
rotations $\{\hat{\R}_{i}\}_{i=1}^N$ by minimizing the squared Frobenius norms:
\begin{equation}\label{eqn:align}
  \bm{\mathrm{S}}^\star=
  \mathop{\arg\min}\limits_{\bm{\mathrm{S}}\in \textrm{SO}(3)}
  \sum\limits_{i=1}^{N}\|\bm{\mathrm{S}}-\R_i^\mathrm{T}\hat{\R}_i\|_\textrm{F}^2
\end{equation}

The solution could be obtained by the existing method \cite{quaternionAvg}.
Then the final loss function is computed as:
\begin{equation}\label{eqn:loss_stage2}
  \mathcal{L}_\mathrm{A} = \frac{1}{N}\sum\limits_{i=1}^N\|\R_i\bm{\mathrm{S}}^\star-\hat{\R}_i\|_2
\end{equation}

The above procedures are fully differentiable, thus the gradient could be 
computed automatically using existing deep-learning libraries. 
It should be pointed out that there is no direct constraint for the predicted 
confidences themselves.
Instead, the only supervision signal for the confidences comes from the 
final predicted absolute rotations, \ie{}, 
the confidences are adjusted automatically 
in order to give a better absolute rotation estimation.
Intuitively, a large confidence score should be assigned to the reliable relative
rotation, and a small confidence score should be assigned to the unreliable relative rotation. 
Please see the analysis in \Cref{subsec:ablation}.

\subsection{Extension to Large-Scale Scenes}\label{subsec:large}

As noted from \cref{fig:architecture}, the input of the network is a 
set of $N$ images at each training step, and both the memory and computational costs
of EAR-Net have to be dependent on the number $N$ accordingly in principle.
With the increase of $N$, the memory and computational cost would 
increase significantly. 

To address the above issue, we generally set the image number $N$ to be a 
small integer (In our low-cost server, $N$ is always set to 7) during 
the training stage, and design the following manner to deal with 
large-scale scenes where hundreds and 
thousands of images are involved at the testing stage.

Specifically, for a testing large-scale scene, 
the feature maps are firstly extracted batch-by-batch from all the testing images 
via the feature encoder in \cref{fig:architecture}, and then they are temporarily stored.
Next, an \textbf{edge-by-edge processing strategy} is used 
to obtain the relative rotations and confidences as:

At each time, only one pair of feature maps, which corresponds to one pair of cameras
that has an edge in the scene graph, is loaded into the GPU memory.
The feature maps are firstly processed by the PFA unit to obtain the pairwise feature vector, then it is 
further decoded into the relative rotation and its confidence. 

As noted from the above edge-by-edge processing strategy, 
since only two feature maps are processed at each time, 
the memory cost would not increase with the total number of images.
Once all relative rotations and the confidences within the scene graph are obtained, 
we employ the CARA module to obtain the final absolute rotations.

\section{Experiments}
\label{sec:experiments}

\subsection{Datasets and Evaluation Metrics}\label{subsec:datasets}

\noindent\textbf{Datasets. }
To verify the effectiveness of the proposed EAR-Net,
we conduct experiments on three public datasets, including
the ScanNet \cite{ScanNet}, the DTU \cite{DTU} and the 7-Scene \cite{7Scene} datasets.
The ScanNet dataset contains 1613 scans collected from 807
indoor scenes, including 1513 training scans and 100 testing scans. 
The DTU dataset contains 22 testing scans
collected in a laboratory setting. 
The images are captured from different angles using a camera mounted on a robot arm.
The 7-Scene dataset is a relatively small dataset 
collected from 7 scenes with 46 scans in total.
Our evaluation is split into three setups, 
namely the basic setup, the large-scale setup, and the cross-dataset setup,
which will be introduced in the following.

\begin{table*}[t]
\fs
\def\len{0.9}
\def\lenn{0.7cm}
\def\lennn{0.9cm}
\def\IRLS{IRLS-$\ell_{\frac{1}{2}}$ \cite{IRLS}}
\def\RAGO{RAGO \cite{RAGO}}
\def\HARA{HARA \cite{HARA}}
\centering
\caption{Comparison on ScanNet \cite{ScanNet}.
  `RR' denotes the relative rotation estimator. `RA' denotes the rotation averaging solver.
  `Mn' denotes the mean error. `Med' denotes the median error.
`MNN' denotes mutual nearest neighbor matching.
`NS/NT' denotes the \textbf{N}umber
of \textbf{S}uccessful/\textbf{T}otal image sets
(same for \Cref{tab:cross}).
The best results are marked in \textbf{bold} face.
}
\label{tab:scannet}
\begin{tabular}{lccccr}
  \toprule
    RR & RA & Mn$\downarrow$ & Med$\downarrow$     & Acc@10\degree$\uparrow$ & NS/NT$\uparrow$\\
  \midrule
  SIFT \cite{SIFT}+MNN                  & \multirow{6}{*}{\IRLS}  & 15.98 & 8.90  & 54.01  & 1475/1485           \\
  SuperPoint \cite{SuperPoint}+MNN      &         & 12.14 & 5.84  & 66.01  & 1483/1485           \\
  LoFTR \cite{LoFTR}                    &         &  7.57 & 4.57  & 82.08  & \textbf{1485}/1485   \\
  ASpanFormer \cite{aspanformer}        &         &  7.06 & 4.30  & 84.41  & 1484/1485            \\
  Reg6D \cite{rot6d}                    &         & 11.38 & 6.35  & 68.34  & \textbf{1485}/1485 \\
  ExtremeRotation \cite{Cai2021extreme} &         &  9.22 & 4.60  & 76.98  & \textbf{1485}/1485 \\
  \midrule                                                         
  SIFT \cite{SIFT}+MNN                  & \multirow{6}{*}{\RAGO}        & 14.65 & 8.32  & 56.22  & \textbf{1485}/1485 \\
  SuperPoint \cite{SuperPoint}+MNN      &         & 12.45 & 6.07  & 65.70  & \textbf{1485}/1485 \\
  LoFTR \cite{LoFTR}                    &         & 7.68  & 4.07  & 81.85  & \textbf{1485}/1485 \\
  ASpanFormer \cite{aspanformer}        &         & 7.11  & 3.74  & 84.02  & \textbf{1485}/1485 \\
  Reg6D \cite{rot6d}                    &         & 11.87 & 6.55  & 67.37  & \textbf{1485}/1485 \\
  ExtremeRotation \cite{Cai2021extreme} &         & 9.83  & 5.04  & 75.26  & \textbf{1485}/1485 \\
  \midrule                                                         
  SIFT \cite{SIFT}+MNN                  & \multirow{6}{*}{\HARA}        & 24.01 & 8.45  & 56.01  & 1447/1485 \\
  SuperPoint \cite{SuperPoint}+MNN      &         & 14.22 & 5.41  & 68.41  & 1479/1485 \\
  LoFTR \cite{LoFTR}                    &         &  7.42 & 4.53  & 83.22  & \textbf{1485}/1485 \\
  ASpanFormer \cite{aspanformer}        &         &  6.91 & 4.26  & 85.17  & 1484/1485 \\
  Reg6D \cite{rot6d}                    &         & 11.06 & 6.09  & 70.57  & \textbf{1485}/1485 \\
  ExtremeRotation \cite{Cai2021extreme} &         &  9.05 & 4.53  & 77.88  & \textbf{1485}/1485 \\
  \midrule
  \multicolumn{2}{l}{PoseDiffusion \cite{wang2023posediffusion}} 
  & 14.44 & 6.17  & 66.50  & \textbf{1485}/1485 \\
  \multicolumn{2}{l}{RelPose++ \cite{lin2024relpose++}} 
  & 9.08 & 3.60 & 83.97 & \textbf{1485}/1485 \\
  \multicolumn{2}{l}{PRAGO \cite{taiana2024prago}}            
  & 8.33 & 4.49 & 78.86  & \textbf{1485}/1485 \\
  DUST3R \cite{wang2024dust3r}  
  & & 4.57     & 3.55   & \textbf{95.39} & \textbf{1485}/1485 \\
  \midrule
  \multicolumn{2}{l}{EAR-Net}  & \textbf{4.03} & \textbf{2.06} & 94.18 & \textbf{1485}/1485 \\
  \bottomrule
\end{tabular}
\end{table*}

\textit{Basic setup:} 
We train and test EAR-Net on ScanNet \cite{ScanNet}.
We follow the official data split for training and testing.
For the training/testing split, we sample 150/15 sets of images with 
size 7 in each scene. This results in 226950/1485 sets of 
images for training/testing
\footnote{Since there are not enough image sets in the scan
  `scene00718\_00', in the test split, only 99 scans are indeed
  used for testing in total. The detailed sampling strategy is provided in the
  Appendix.}.
  
\textit{Large-scale setup: } 
To further evaluate the performance of EAR-Net on large-scale scenes, 
we conduct experiments on the full scans of ScanNet. 
Specifically, we evaluate the model on all the scans which contain 
more than 3000 images. This results in 22 large-scale scenes in total.

\textit{Cross-dataset setup:} 
We further evaluate the generalization abilities on
the DTU \cite{DTU} and 7-Scene datasets \cite{7Scene}.
We train the model on ScanNet and evaluate on the other two.
For the DTU dataset, we sample 100 image sets of size 7
from each scan, resulting in 2200 image sets for testing. 
For the relatively small 7-Scene dataset, 
we evaluate on the full set. We sample 20 image sets with size 7
from each scan, resulting in 920 image sets in total.

\noindent\textbf{Evaluation Metrics. }
For each image set, we first obtain the
transformation $\bm{\mathrm{S}}^\star$ from the estimated
absolute rotations to the ground-truth by minimizing the above
objective function in \Cref{eqn:align}. Then, for each
estimated global rotation matrix $\R$ and the
corresponding ground-truth $\hat{\R}$, we compute the rotation error:
$\arccos((\mathrm{tr}(\hat{\R}^\mathrm{T}\R\bm{\mathrm{S}}^\star)-1)/2)$,
where `$\mathrm{tr}(\cdot)$' denotes the trace. Finally, the following
metrics are reported according to the rotation error:
the mean error, the median error as done in \cite{hartleyAvg,IRLS,HARA}.
Moreover, we also report the percentage of rotation error
that is under $10^\circ$.

\subsection{Implementation Details}\label{subsec:implementation}
EAR-Net is implemented in PyTorch. Firstly, the feature 
encoder, the PFA unit and the rotation branch of the dual-branch decoder are 
pretrained for 30 epochs with a learning rate of $5\times 10^{-4}$ and a 
batch size of 20 using the Adam optimizer. Then, we initialize the 
confidence branch of the decoder so that it outputs a constant 
value of 0.5 for all pairwise rotations. This could be achieved by 
dividing the logits (before the \texttt{sigmoid} activation in the last layer 
of the confidence branch) 
by a large constant. Next, the whole model is trained with a learning rate of 
$1\times 10^{-4}$ and a batch size of 8 using the Adam optimizer. 
We perform $T=3$ iterations of confidence-aware optimization.
In total, the model is trained for 50 epochs.

\subsection{Comparative Evaluation Under Basic Setup}\label{subsec:evaluation}

In this setup, we train and test the proposed EAR-Net on ScanNet \cite{ScanNet}.
We compare EAR-Net with some typical methods as listed in \cref{tab:scannet}.
It has to be indicated that DUST3R \cite{wang2024dust3r} is a large foundation 
model trained on eight large-scale datasets 
(including ScanNet++ \cite{yeshwanth2023scannet++}, 
which has a distribution close to that of ScanNet and contains more video scans), 
while the other comparative methods are only trained on ScanNet. 
Additionally, for the first six approaches, 
we adopt the following approaches for relative rotation estimation:
1) two detector-matcher-based methods with the 5-point algorithm
\cite{nister2004efficient}: 
SIFT \cite{SIFT}-MNN (mutual nearest neighbor matcher),
SuperPoint \cite{SuperPoint}-MNN;
2) two end-to-end matcher-based methods LoFTR \cite{LoFTR}, ASpanFormer \cite{aspanformer}
with the 5-point algorithm;
3) two end-to-end relative rotation estimation methods Reg6D \cite{rot6d}
and ExtremeRotation \cite{Cai2021extreme}.
Then, once the relative rotations are obtained, the three state-of-the-art solvers
(IRLS-$\ell_\frac{1}{2}$ \cite{IRLS}, HARA \cite{HARA}, and
the learning-based RAGO \cite{RAGO}) for rotation averaging are implemented respectively 
to calculate the absolute rotations.

\begin{figure}[t]
    \centering
    \includegraphics[width=\linewidth]{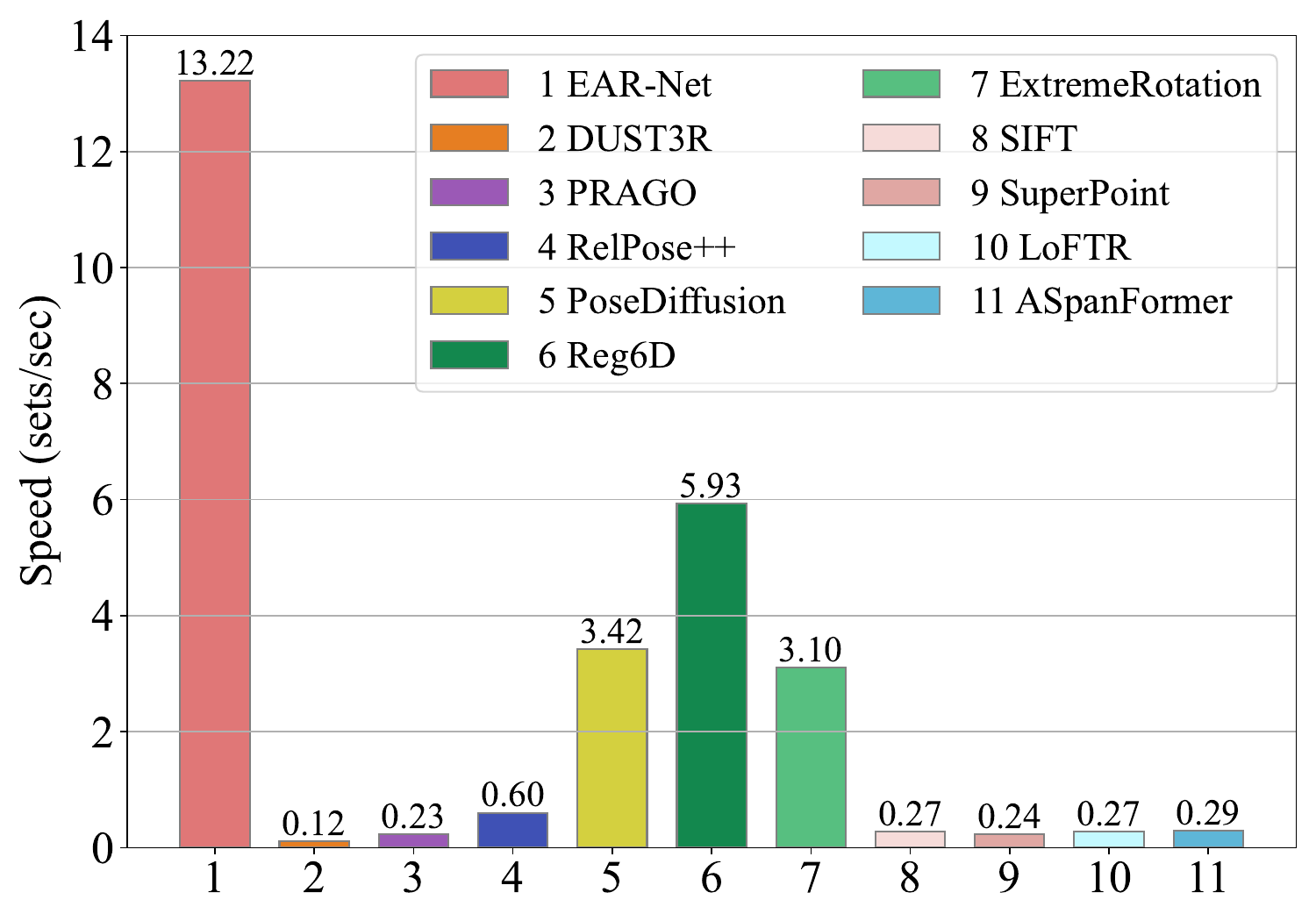}
    \caption{
        Comparison of inference speed on the ScanNet dataset \cite{ScanNet}.
        All the methods are tested on an RTX 2080Ti GPU with batch size 1.
      }
    \label{fig:speed}
\end{figure}

\begin{table}[t]
\fs
\centering
\begin{tabular}{lccc}
\toprule
Method               & Model (s) & Averaging (s) & Total (s) \\
\midrule
SIFT+HARA            & 3.698 & 0.006 & 3.704 \\
SuperPoint+HARA      & 4.161 & 0.006 & 4.167 \\
LoFTR+HARA           & 3.698 & 0.006 & 3.704 \\
ASpanFormer+HARA     & 3.442 & 0.006 & 3.448 \\
Reg6D+HARA           & 0.163 & 0.006 & 0.169 \\
ExtremeRotation+HARA & 0.317 & 0.006 & 0.323 \\
\midrule
EAR-Net              & \bf 0.072 & \bf 0.004 & \bf 0.076 \\
\bottomrule
\end{tabular}
\vspace{-0.1in}
\caption{Comparison of runtime (in seconds) spent in the model part, the averaging part, and the whole method (Total).}
\label{tab:time}
\end{table}

\Cref{tab:scannet} lists the numbers of the successfully recovered
rotations from the total 1485 testing image sets from ScanNet \cite{ScanNet}
by all the methods in its right-most column, and reports the
corresponding results obtained from these successful images.
As seen from this table, the comparative methods
\cite{SIFT,SuperPoint,SuperGlue,aspanformer,LoFTR,patch2pix} sometimes fail
to register all image sets, possibly due to (1) an insufficient number of
inlier matches and (2) the outlier relative rotation
filtering step in HARA \cite{HARA}.
EAR-Net does not only recover all the testing images successfully 
but also outperforms all the non-pre-trained comparative methods significantly under
all the five metrics. 
Moreover, even compared with the pre-trained large model DUST3R \cite{wang2024dust3r}, 
the proposed EAR-Net performs slightly worse under the metric `accuracy under 10\degree', 
but much better under the rest of the metrics. 
All the above results demonstrate the effectiveness of the proposed EAR-Net.

\noindent\textbf{Inference Speed. }
Here we further evaluate the effectiveness of EAR-Net in terms of inference speed. 
The inference speeds by the proposed method and the comparative methods
\cite{SIFT,SuperPoint,aspanformer,LoFTR,rot6d} with the HARA solver \cite{HARA} 
(which demonstrates better performance than the other solvers \cite{IRLS,RAGO} in most cases) 
on ScanNet \cite{ScanNet} are shown in \Cref{fig:speed}.
All the referred methods are evaluated on a server with an RTX 2080Ti GPU with 
a batch size of 1.
As seen from this figure, EAR-Net runs at a speed of \textbf{13.2} image sets per second, which 
is significantly faster than all these comparative methods. 
Note that unlike the existing 5-point algorithm based methods 
\cite{SIFT,SuperPoint,LoFTR,aspanformer}, 
our end-to-end EAR-Net is GPU-friendly, and its inference speed could be further improved 
by parallel computing with a larger batch size. For example, with a batch size of 30, EAR-Net runs at 
around \textbf{192.8} image sets per second. In this case, compared with the 5-point algorithm 
based methods (SIFT, SuperPoint, LoFTR, ASpanFormer), EAR-Net is more than $\mathbf{640\times}$ faster.

\begin{table*}[t]
  \centering
  \fs
  \def\len{0.85cm}
  \def\ASPAN{\multicolumn{3}{|c}{ASpan+HARA}}
  \def\REG{\multicolumn{3}{|c}{Reg6D+HARA}}
  \def\ER{\multicolumn{3}{|c}{ER+HARA}}
  \def\EAR{\multicolumn{3}{|c}{EAR-Net}}
  \def\data{\multicolumn{2}{c}{Dataset}}
  \begin{tabular}{cC{\len}|rR{0.2in}C{0.24in}|rR{0.2in}C{0.25in}|rR{0.2in}C{0.25in}|rrC{0.25in}}
    \toprule
    \data & \ASPAN & \REG & \ER & \EAR                                          \\
    \midrule
    Scene & \#Cam. & Mn & Med & t & Mn & Med & t & Mn & Med & t & Mn & Med & t  \\
    0711\_00 & 3361 &  54.1 & 43.6 & \textcolor{red}{8.1h}  & 27.5 & 25.5 & 1386s & 37.1 & 35.9 & 1893s & \textbf{14.8} & \textbf{14.7} & \textbf{717s}  \\
    0712\_00 & 4787 &  68.9 & 59.4 & \textcolor{red}{11.2h} & 58.3 & 58.8 & 2222s & 58.4 & 63.9 & 2762s &  \textbf{8.4} &  \textbf{6.9} & \textbf{1023s} \\
    0721\_00 & 3773 &  59.6 & 61.5 & \textcolor{red}{10.4h} & 31.5 & 32.7 & 1543s & 40.6 & 33.1 & 1989s & \textbf{18.3} & \textbf{15.0} &  \textbf{795s} \\
    0736\_00 & 8009 &  72.3 & 61.5 & \textcolor{red}{21.9h} & 89.8 & 90.9 & 3215s & 84.4 & 80.9 & \textcolor{red}{1.3h} & \textbf{58.2} & \textbf{54.5} & \textbf{1743s} \\
    0737\_00 & 3076 &  73.9 & 67.4 &  \textcolor{red}{8.0h} & 51.7 & 54.6 & 1157s & 79.6 & 81.3 & 1839s & \textbf{20.2} & \textbf{14.4} &  \textbf{657s} \\
    0739\_00 & 4449 &  41.7 & 36.1 & \textcolor{red}{11.1h} & 42.9 & 39.5 & 1708s & 44.3 & 44.5 & 2832s & \textbf{11.4} & \textbf{11.8} & \textbf{1001s} \\
    0744\_00 & 3127 &  63.0 & 56.7 &  \textcolor{red}{7.2h} & 32.6 & 21.2 & 1223s & 17.0 & 14.4 & 1694s &  \textbf{6.0} &  \textbf{5.8} &  \textbf{671s} \\
    0747\_00 & 5024 &  101.2& 103.0& \textcolor{red}{14.1h} & 80.5 & 56.5 & 1895s & 80.4 & 60.4 & 2911s & \textbf{28.9} & \textbf{32.3} & \textbf{1094s} \\
    0752\_00 & 3050 &  69.4 & 56.5 &  \textcolor{red}{7.6h} & 35.3 & 32.2 & 1121s & 21.9 & 20.9 & 1678s & \textbf{12.6} & \textbf{13.0} &  \textbf{696s} \\
    0753\_00 & 3389 &  60.1 & 57.4 &  \textcolor{red}{8.1h} & 23.2 & 16.6 & 1248s & 33.1 & 31.2 & 1846s & \textbf{11.7} & \textbf{10.4} &  \textbf{770s} \\
    0754\_00 & 3218 &  29.5 & 25.7 &  \textcolor{red}{7.4h} & 15.9 & 16.7 & 1193s & 14.0 & 14.0 & 1713s & \textbf{12.4} & \textbf{12.4} &  \textbf{743s} \\
    0755\_00 & 3546 &  49.3 & 49.1 &  \textcolor{red}{9.6h} & 26.0 & 26.3 & 1309s & 31.8 & 34.1 & 1898s & \textbf{11.5} &  \textbf{6.6} &  \textbf{872s} \\
    0756\_00 & 3503 &  36.7 & 34.9 &  \textcolor{red}{8.1h} & 20.6 & 14.5 & 1324s & 22.3 & 14.3 & 1828s & \textbf{10.8} & \textbf{10.2} &  \textbf{811s} \\
    0757\_00 & 8336 &  40.6 & 37.2 & \textcolor{red}{21.6h} & 58.2 & 61.6 & 3185s & 84.2 & 78.0 & \textcolor{red}{1.2h} & \textbf{10.4} &  \textbf{9.0} & \textbf{1972s} \\
    0761\_00 & 5190 &  58.7 & 51.9 & \textcolor{red}{13.7h} & 33.9 & 26.3 & 2010s & 48.8 & 43.4 & 2657s &  \textbf{9.3} &  \textbf{9.0} & \textbf{1200s} \\
    0766\_00 & 3504 &  58.6 & 57.7 &  \textcolor{red}{9.7h} & 40.1 & 38.2 & 1326s & 66.0 & 27.9 & 1784s & \textbf{28.5} & \textbf{26.9} &  \textbf{793s} \\
    0768\_00 & 4026 &  74.7 & 71.5 &  \textcolor{red}{9.0h} & 53.8 & 52.3 & 1510s & 25.2 & 26.8 & 2054s & \textbf{10.6} & \textbf{10.7} &  \textbf{884s} \\
    0770\_00 & 3414 &  62.3 & 48.1 &  \textcolor{red}{8.6h} & 40.8 & 43.3 & 1345s & 33.0 & 28.9 & 1746s &  \textbf{9.8} &  \textbf{7.0} &  \textbf{791s} \\
    0776\_00 & 3478 &  60.4 & 54.3 &  \textcolor{red}{8.6h} & 45.2 & 46.1 & 1338s & 27.2 & 30.1 & 1782s &  \textbf{4.8} &  \textbf{4.3} &  \textbf{836s} \\
    0784\_00 & 4926 &  60.7 & 55.8 & \textcolor{red}{12.5h} & 39.4 & 40.3 & 1876s & 49.2 & 44.4 & 2732s & \textbf{16.1} & \textbf{17.9} & \textbf{1108s} \\
    0785\_00 & 3980 &  47.6 & 46.1 & \textcolor{red}{10.9h} & 20.3 & 18.0 & 1558s & 36.4 & 36.7 & 2168s & \textbf{10.4} &  \textbf{8.6} &  \textbf{876s} \\
    0793\_00 & 3457 &  72.0 & 53.5 &  \textcolor{red}{8.0h} & 58.1 & 58.9 & 1399s & 17.8 & 14.0 & 1835s & \textbf{10.0} & \textbf{10.2} &  \textbf{756s} \\
    \bottomrule
  \end{tabular}
  \caption{Comparison on large-scale scenes from the ScanNet dataset \cite{ScanNet}. 
  `Mn' denotes the mean error. `Med' denotes the median error. `t' denotes the time cost.
  `ER' denotes the ExtremeRotation \cite{Cai2021extreme}.
  `ASpan' denotes the ASpanFormer matcher \cite{aspanformer}.
  The best results are marked in \textbf{bold} face.
  The time costs that are longer than one hour are marked in \textcolor{red}{red}.
  }
  \label{tab:large_graphs}
\end{table*}

Moreover, for the proposed method and the comparative methods 
\cite{SIFT,SuperPoint,LoFTR,aspanformer,rot6d,Cai2021extreme,HARA} 
that could be simply divided into the model part and the averaging part, 
we report the time spent in these parts separately in \cref{tab:time}. 
As seen from this table, all the involved methods spend significantly less time on the 
averaging part than on the model part, mainly because their model parts have to process 
high-dimensional image data, while their averaging parts only 
process small-sized 3-order rotation matrices. 
Moreover, the runtime of the model part of the proposed EAR-Net is 
much lower than those of the other methods,
mainly because the proposed method is an end-to-end method 
and its model part only performs a single feedforward computation. 
The runtime of the averaging part of EAR-Net is also lower than those of the other methods, 
mainly because the HARA solver \cite{HARA} requires an additional preprocessing step to 
filter out relative rotation outliers.

\begin{figure*}[!h]
    \centering
    \includegraphics[width=1.0\linewidth]{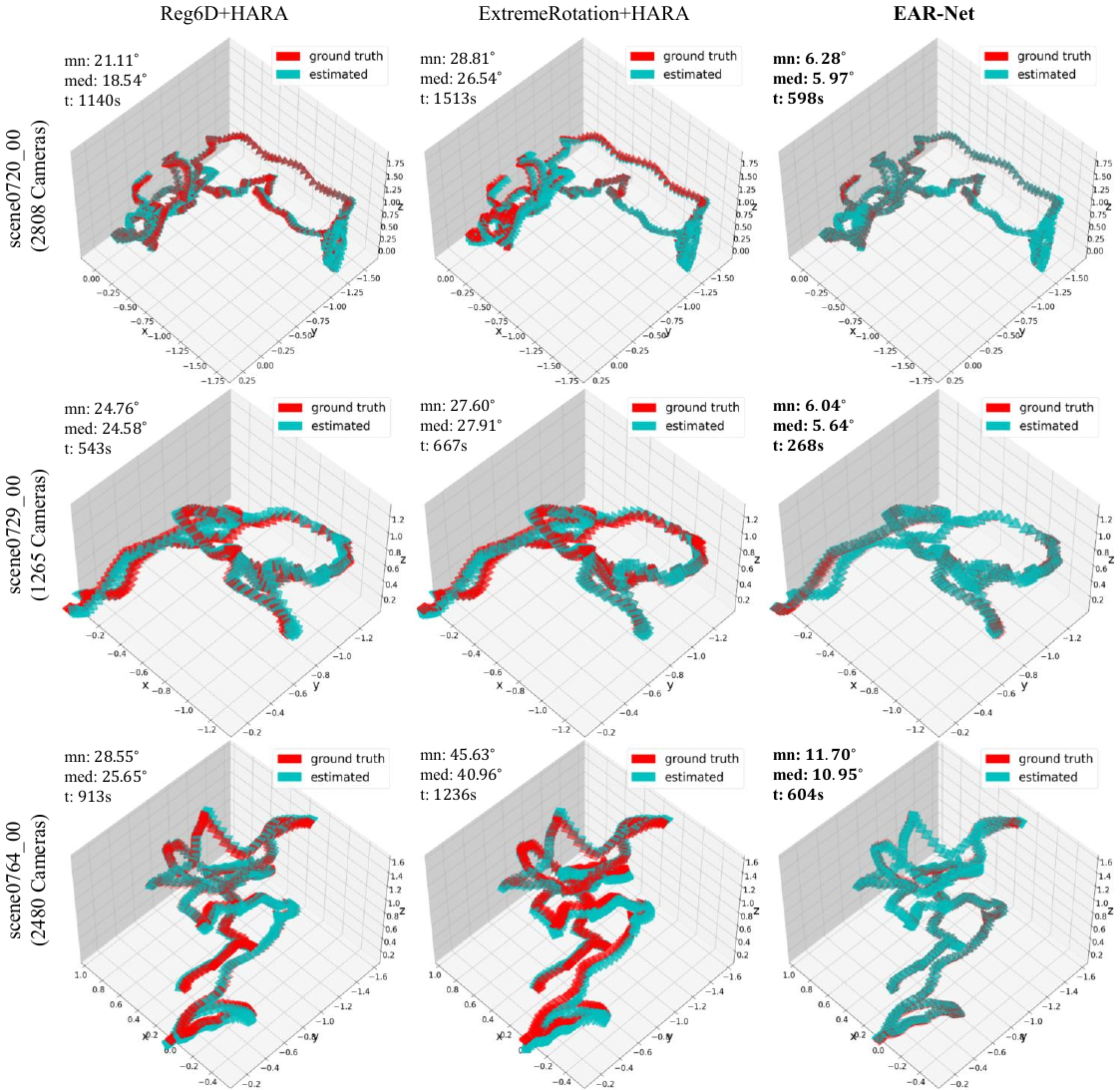}
    \caption{Qualitative comparison of the estimated absolute rotations on ScanNet. 
    For a clear comparison, both the estimated rotations and the ground-truth rotations 
    are downsampled by 5 in all the sub-figures. In each sub-figure, 
    each pair of rectangular pyramids (\textcolor{cyan}{cyan} and \textcolor{red}{red}), 
    which share the same apex, represent the estimated rotation and ground-truth rotation respectively, 
    and their apex represents the ground-truth position of the corresponding camera. 
    The three sub-figures in the first column show the results of Reg6D\cite{rot6d}+HARA\cite{HARA} 
    on three scene sequences, the corresponding sub-figures in the second column 
    show the results of ExtremeRotation\cite{Cai2021extreme}+HARA\cite{HARA}, 
    while the corresponding sub-figures in the third column show the results of the proposed method.  
    Additionally, in the top left corner of each sub-figure, the computed mean error, 
    median error, and the computational time (second) of the corresponding method are reported. 
    }
    \label{fig:vis}
\end{figure*}

\subsection{Comparative Evaluation Under Large-Scale Setup}\label{subsec:evaluation_large}

In this subsection, we conduct experiments on the 22 large-scale scenes from ScanNet \cite{ScanNet}. 
It is noted that the four 5-point algorithm based methods (SIFT, SuperPoint, LoFTR, ASpanFormer) 
are very slow to evaluate on such large-scale scenes. 
And among them, ASpanFormer performs much better than the other three 5-point algorithm based methods.
Thus, in these methods, we only evaluate the performance of ASpanFormer \cite{aspanformer}. 
In addition, we also evaluate the performance of the regression-based Reg6D \cite{rot6d} 
and ExtremeRotation \cite{Cai2021extreme}.
Then, we obtain the absolute rotations with the HARA rotation averaging solver, which demonstrates 
better performance than the other solvers \cite{IRLS,RAGO} in most cases. 
The mean errors, the median errors, and the time costs of 
these comparative methods in each scene are reported in \cref{tab:large_graphs}.
As seen from this table, the 5-point algorithm based 
method ASpanFormer \cite{aspanformer} performs 
poorer than the regression-based methods Reg6D and ExtremeRotation.
This is mainly because the baselines between pairs of cameras 
are much wider in the large-scale scenes than those in the basic setup,
so that it is difficult for ASpanFormer to obtain high-accuracy matching points, 
however, the Reg6D and ExtremeRotation are end-to-end learning-based methods  
that do not explicitly extract feature points, as indicated in \cite{chen2021wide,Cai2021extreme}. 
Moreover, the proposed EAR-Net outperforms the other three 
comparative methods \cite{Cai2021extreme,IRLS,HARA,RAGO} 
significantly on all of the evaluated scenes.
It is also noted that the EAR-Net has a significantly lower 
inference time cost in all the evaluated scenes, especially compared to the 
5-point algorithm based ASpanFormer \cite{aspanformer} which generally takes 7-20 hours.
The above experimental results demonstrate the effectiveness of 
our method on large-scale scenes in terms of both accuracy and inference speed.

\begin{table*}[t]
    \fs
  \def\len{1.1cm}
  \def\lenn{0.65cm}
  \def\lennn{1.5cm}
  \def\IRLS{IRLS-$\ell_{\frac{1}{2}}$}
  \def\RAGO{RAGO}
  \def\HARA{HARA}
    \centering
      \caption{Cross-dataset comparison on the DTU \cite{DTU} and 7-Scene \cite{7Scene} datasets. 
      The models are trained on the ScanNet dataset.
      The best results are marked in \textbf{bold} face.}
      \label{tab:cross}
\begin{tabular}{lcC{\lenn}C{\lenn}C{\len}R{\lennn}@{\hskip 0.3in}C{\lenn}C{\lenn}C{\len}R{\lennn}}
  \toprule
  \multirow{2}{*}{RR} & \multirow{2}{*}{RA} & \multicolumn{4}{c}{\textbf{DTU}} &  \multicolumn{4}{c}{\textbf{7-Scene}} \\
  \cmidrule(lr{16pt}){3-6} \cmidrule(lr){7-10}
  &&Mn$\downarrow$ & Med$\downarrow$ & Acc@10$^\circ$$\uparrow$ & NS/NT$\uparrow$ &
    Mn$\downarrow$ & Med$\downarrow$ & Acc@10$^\circ$$\uparrow$ & NS/NT$\uparrow$\\
  \midrule
  SIFT+MNN          & \multirow{6}{*}{\IRLS}       & 24.93 & 21.92 & 17.52 & \textbf{2200}/2200 &  9.61 & 5.26 & 71.69 & \textbf{920}/920\\
  SuperPoint+MNN    &        & 19.76 & 16.11 & 32.79 & \textbf{2200}/2200 &  8.06 & 3.86 & 79.67 & \textbf{920}/920\\
  LoFTR             &        & 18.05 & 14.89 & 27.80 & 2195/2200          &  6.79 & 5.06 & 84.47 & \textbf{920}/920\\
  ASpanFormer       &        & 16.47 & 13.44 & 32.99 & 2195/2200          &  6.06 & 4.79 & 87.28 & \textbf{920}/920\\
  Reg6D             &        & 22.45 & 19.55 & 18.49 & \textbf{2200}/2200 & 11.12 & 7.60 & 63.65 & \textbf{920}/920\\
  ExtremeRotation   &        & 21.82 & 18.01 & 21.33 & \textbf{2200}/2200 &  9.24 & 6.58 & 70.82 & \textbf{920}/920\\
  \midrule                                                                                      
  SIFT+MNN          &  \multirow{6}{*}{\RAGO}      & 22.23 & 18.71 & 21.68 & \textbf{2200}/2200 & 11.07 & 7.08 & 63.87 & \textbf{920}/920\\
  SuperPoint+MNN    &        & 18.91 & 14.63 & 33.05 & \textbf{2200}/2200 &  9.38 & 5.61 & 72.41 & \textbf{920}/920\\
  LoFTR             &        & 17.33 & 13.81 & 33.21 & \textbf{2200}/2200 &  7.15 & 4.75 & 82.66 & \textbf{920}/920\\
  ASpanFormer       &        & 14.03 & 10.86 & 45.49 & \textbf{2200}/2200 &  6.24 & 4.41 & 86.29 & \textbf{920}/920\\
  Reg6D             &        & 22.72 & 19.42 & 19.52 & \textbf{2200}/2200 & 11.69 & 7.86 & 62.10 & \textbf{920}/920\\
  ExtremeRotation   &        & 21.87 & 17.86 & 21.76 & \textbf{2200}/2200 &  9.90 & 6.86 & 68.42 & \textbf{920}/920\\
  \midrule                                                                                      
  SIFT+MNN          & \multirow{6}{*}{\HARA}  & 36.44 & 21.39 & 21.55 & 2164/2200          & 11.95 & 4.93 & 72.66 & \textbf{920}/920\\
  SuperPoin+MNN     &        & 20.03 & 13.76 & 38.90 & 2193/2200          &  8.19 & 3.65 & 82.70 & \textbf{920}/920\\
  LoFTR             &        & 18.18 & 14.53 & 29.17 & 2194/2200          &  6.57 & 5.00 & 85.73 & \textbf{920}/920\\
  ASpanFormer       &        & 17.12 & 13.62 & 32.72 & 2191/2200          &  6.02 & 4.79 & 87.73 & \textbf{920}/920\\
  Reg6D             &        & 22.62 & 19.50 & 18.74 & \textbf{2200}/2200 & 11.00 & 7.49 & 65.00 & \textbf{920}/920\\
  ExtremeRotation   &        & 21.88 & 17.82 & 21.88 & \textbf{2200}/2200 &  9.33 & 6.48 & 71.30 & \textbf{920}/920\\
  \midrule
  \multicolumn{2}{l}{PoseDiffusion} 
  & 27.04  & 21.52  & 16.69 & \textbf{2200}/2200 
  & 16.36  & 8.00   & 58.07 & \textbf{920}/920 \\
  \multicolumn{2}{l}{RelPose++} 
  & 23.66 & 17.87 & 23.00 & \textbf{2200}/2200 
  & 10.48 & 4.94  & 77.25 & \textbf{920}/920 \\
  \multicolumn{2}{l}{PRAGO}            
  & 15.77  & 12.92 & 36.94 & \textbf{2200}/2200
  & 6.42   & 4.61  & 85.45  & \textbf{920}/920 \\
  \midrule
  \multicolumn{2}{l}{EAR-Net} & \textbf{13.81} & \textbf{10.68} & \textbf{46.64}   & \textbf{2200}/2200 
                              & \textbf{4.43}  & \textbf{3.26}  & \textbf{94.47}   & \textbf{920}/920\\
  \bottomrule
\end{tabular}
\end{table*}

Besides the above quantitative results, 
we also visualize the estimated rotations on several scenes with 1000-3000 images from ScanNet, 
including `scene720\_00', `scene729\_00', `scene764\_00'. 
The nine sub-figures in Figure 6 show the visualization results of the proposed method and 
two state-of-the-art methods (Reg6D+HARA and ExtremeRotation+HARA) on the three scenes respectively. 
In addition, in the top left corner of each sub-figure, the computed mean error, median error, 
and the computational time (second) of the corresponding method are also reported. 
As seen from \cref{fig:vis}, the estimated absolute rotations by the proposed method align 
significantly better with the ground-truths than those of the two comparative methods, 
and the proposed method also runs much faster than its two counterparts.

\subsection{Comparative Evaluation Under Cross-Dataset Setup}\label{subsec:evaluation_cross}

In this subsection, we evaluate EAR-Net and the comparative methods under the cross-dataset setup.
Specifically, all the non-pre-trained methods that are trained on
ScanNet (except SuperPoint we use the released model that is trained on CoCo \cite{coco} by the author)
are further evaluated on DTU \cite{DTU} and 7-Scene \cite{7Scene}, 
and the corresponding results are reported in \Cref{tab:cross}.
As seen from this table, EAR-Net also performs best among all the referred methods,
consistent with the results in \Cref{tab:scannet}. These results demonstrate the
effectiveness of the proposed EAR-Net under the cross-dataset setup.

\begin{table}
  \centering
  \fs
  \def\len{0.8cm}
  \begin{tabular}{lccc}
    \toprule
    Method & Mn$\downarrow$ & Med$\downarrow$ & Acc@10\degree$\uparrow$\\
    \midrule
    w/o end2end     & 9.08 & 3.92 & 75.75 \\
    w/o pretraining & 6.05 & 3.09 & 88.25 \\
    w/o confidence  & 9.31 & 3.85 & 78.71 \\
    w/o CAI         & 4.83 & 2.09  & 92.69 \\
    \midrule
    Full            & \textbf{4.03} & \textbf{2.06} & \textbf{94.18} \\
    \bottomrule
  \end{tabular}
  \caption{Ablation study on ScanNet \cite{ScanNet}.
          `Mn' and `Med' denote the mean and median error respectively.
  }
  \label{tab:ablation_scannet}
\end{table}

\subsection{Ablation Study}\label{subsec:ablation}

This subsection provides ablation studies on ScanNet \cite{ScanNet}
to evaluate the effect of the following key components:

\noindent\textbf{Effect of End-to-End Training. }
Here, the model is only trained to output relative rotations, and then
the absolute rotations are obtained via the confidence-aware optimization algorithm 
by weighting all edges equally (w/o end2end). 
As seen from \Cref{tab:ablation_scannet},
`w/o end2end' causes the mean and median errors to increase by
\textbf{125.3\%}(=9.08/4.03-1) and \textbf{90.3\%}(=3.92/2.06-1) respectively, 
which indicates our model benefits a lot from end-to-end training.

\noindent\textbf{Effect of Pretraining. }
The performance of EAR-Net without pretraining the feature encoder
and rotation branch is reported in \Cref{tab:ablation_scannet} (w/o pretraining). 
As seen from this table, `w/o pretraining' has a large negative 
impact on the final performance.
This is because the feature encoder and rotation branch
are not initialized well, making the end-to-end training converge
to a worse local minimum.
This observation is consistent with other end-to-end learning methods in other 
visual tasks \cite{dsac,yi2016lift}.

\begin{figure}
    \def\sca{0.9}
    \centering
    \subfigure[Ablation of the confidence.]{\includegraphics[width=\sca\linewidth,clip]{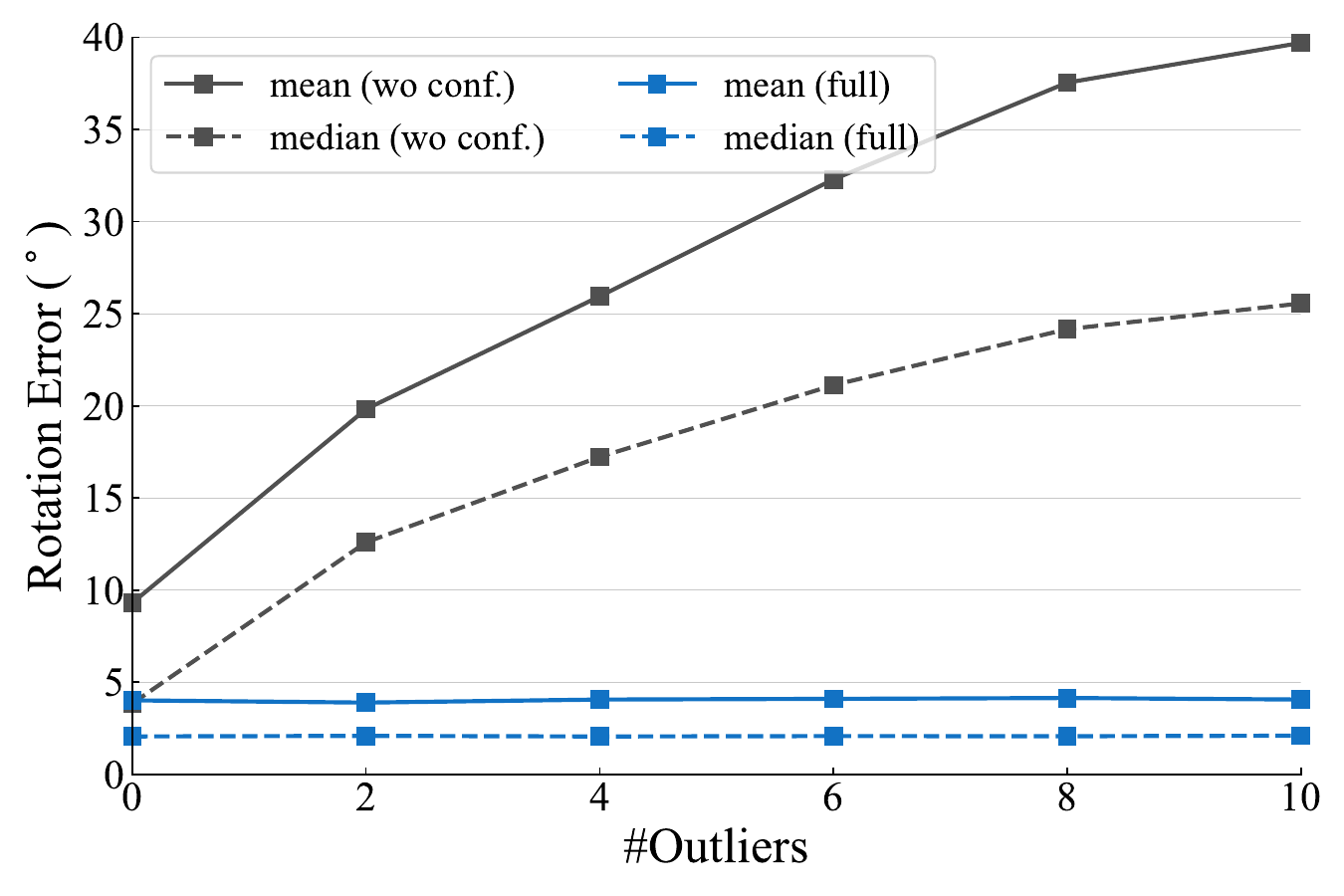}\label{fig:conf_noise}}
    \hfill
    \subfigure[Ablation of the CAI approach.]{\includegraphics[width=\sca\linewidth,clip]{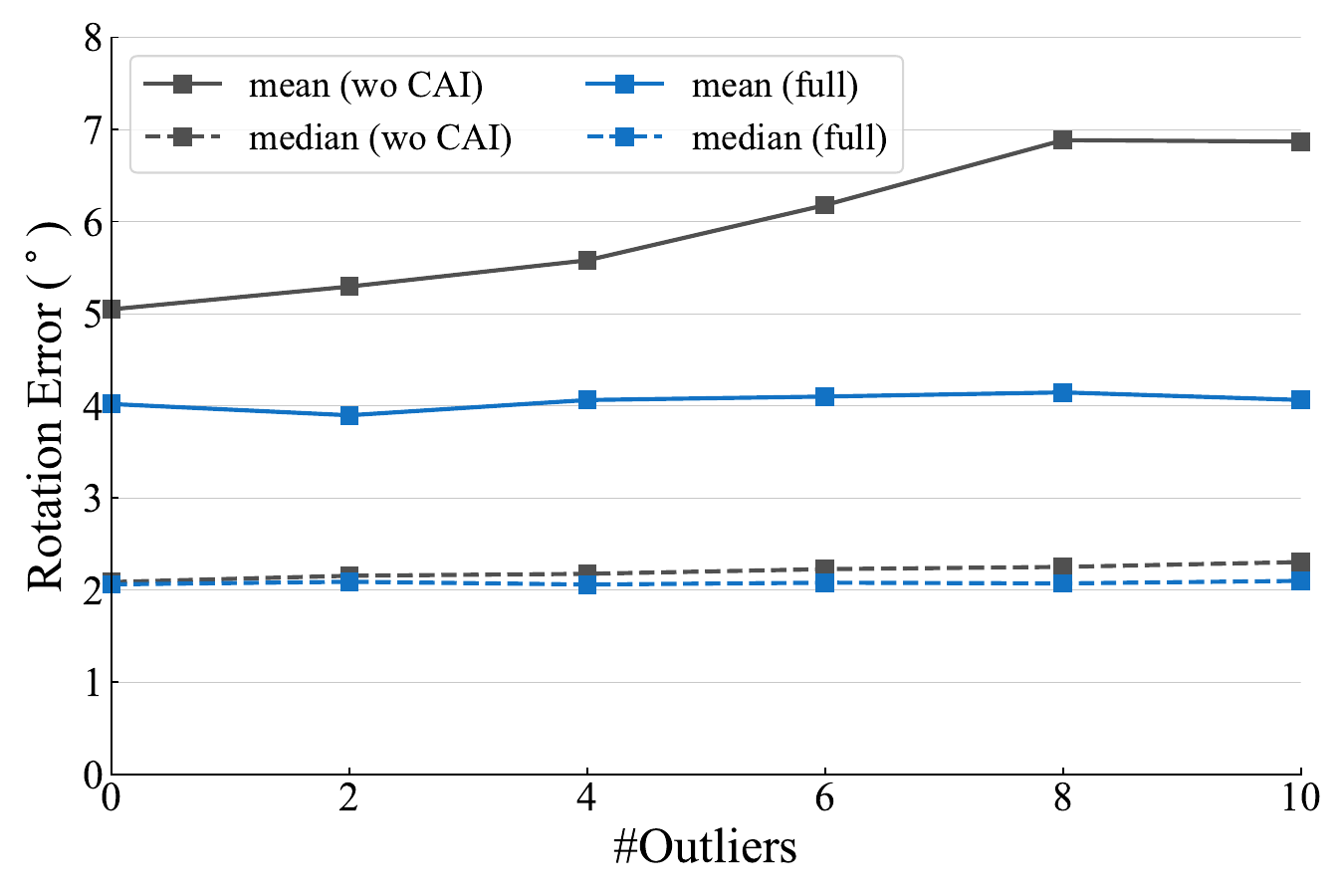}\label{fig:CAI}}
    \vspace{-0.1in}
  \caption{Ablation study on corrupted data. 
  The image sets are corrupted with different numbers of outlier images that have 
  no overlap with the remaining ones.}
    \label{fig:corrupted}
\end{figure}

\begin{figure}[t]
  \centering
  \includegraphics[width=\imsc\linewidth,clip]{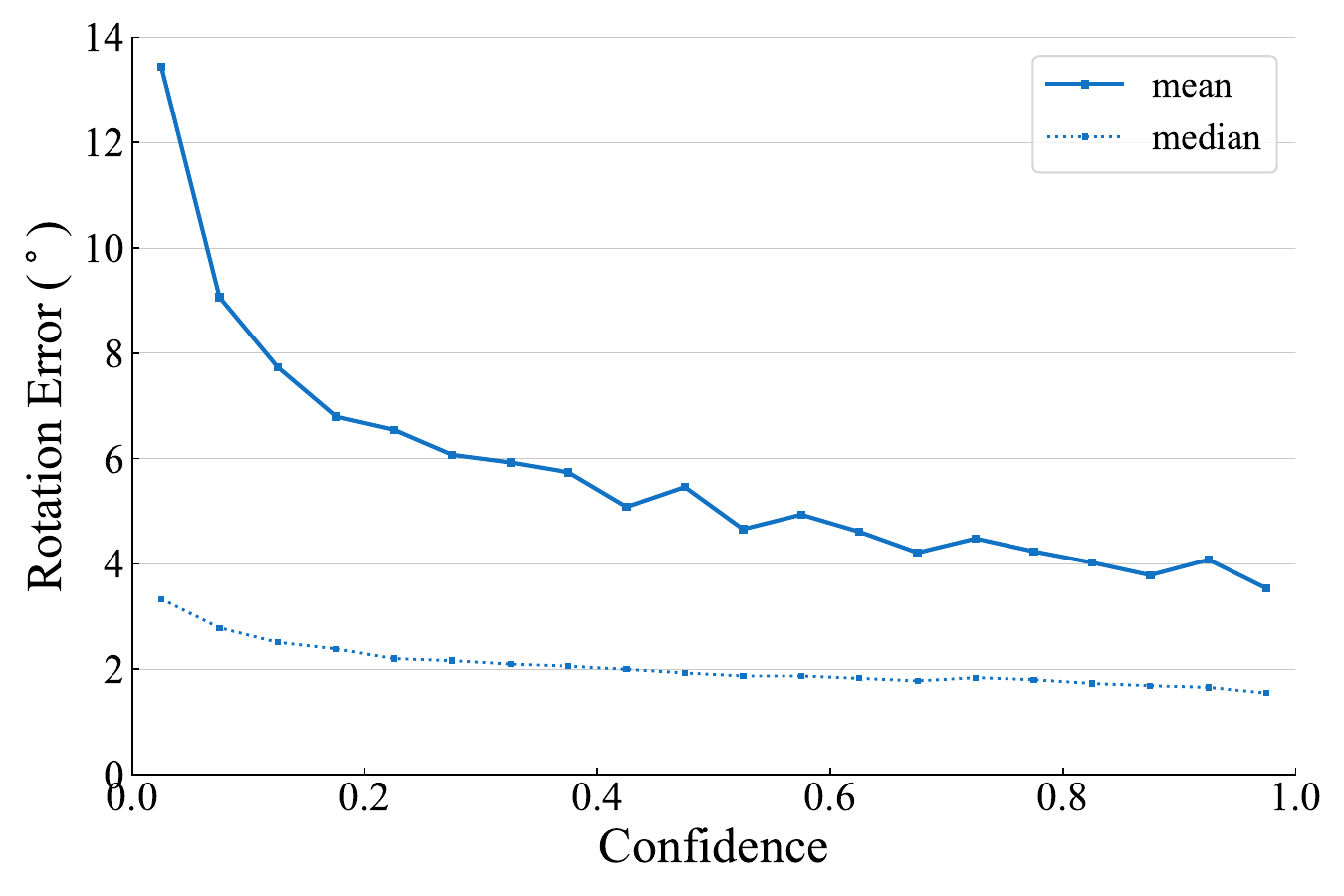}
  \vspace{-0.1in}
  \caption{Relationship of the relative rotation error and confidence
    on ScanNet \cite{ScanNet}. The confidence decoder tends to predict small/large 
    confidences on relative rotations with large/small errors.
  }
  \label{fig:err_conf}
\end{figure}

\begin{figure}[t]
  \centering
 \includegraphics[width=0.8\linewidth,clip]{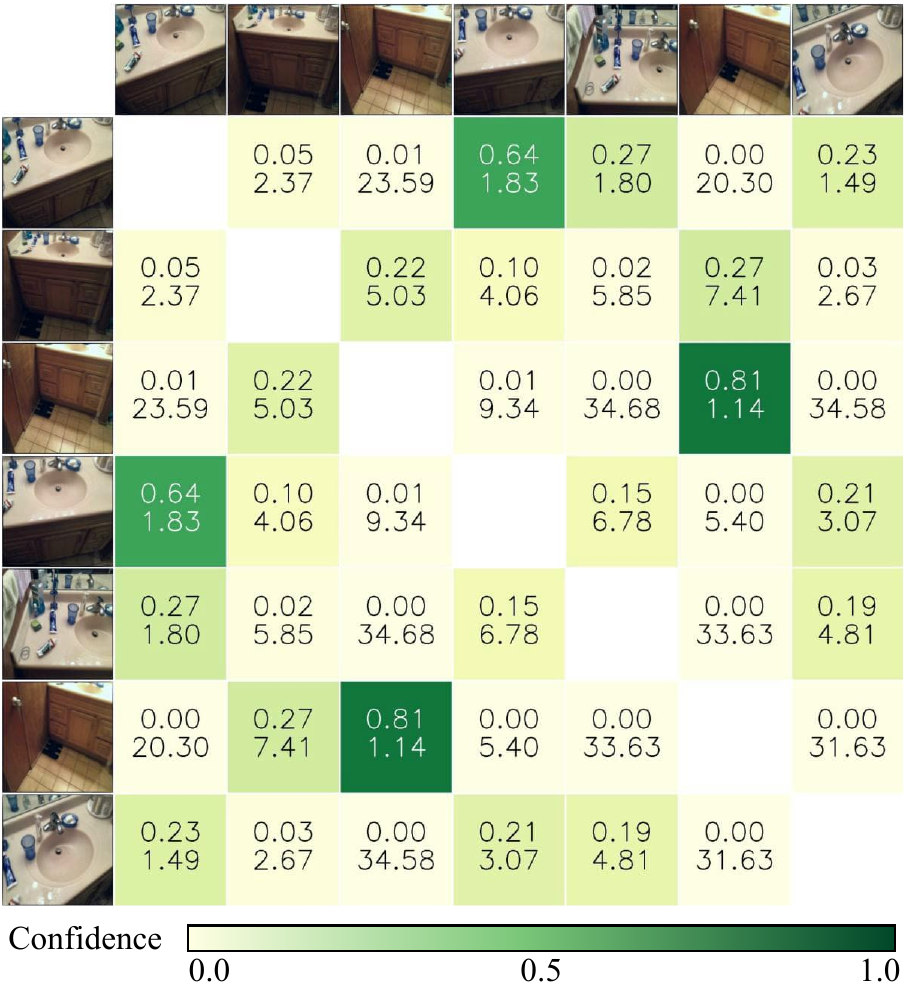}
 \vspace{-0.1in}
  \caption{Visualization of the pairwise relative rotation errors and their 
  corresponding confidences predicted by the dual-branch decoder 
  on the ScanNet dataset \cite{ScanNet}.
    The numbers denote the confidence (top) and the error (bottom, in degree).
  }
  \label{fig:weight_viz}
\end{figure}

\noindent\textbf{Effect of Confidence. }
To evaluate the effect of the learned confidence, the confidence branch
is removed, and all edges are weighted equally
(w/o confidence). 

Firstly, as seen from \Cref{tab:ablation_scannet},
`w/o confidence' causes the mean and median errors to increase
by \textbf{131.0\%} and \textbf{86.9\%} respectively, which demonstrates that the
learned confidence is important for improving the performance of absolute rotation estimation.

Secondly, we conduct
the following experiments to demonstrate the robustness of EAR-Net due to the
learned confidence: 
For each sampled image set with size 7 on ScanNet, we corrupt them  by appending another
$\{0,2,4,6,8,10\}$ randomly selected images from other scenes with no overlap to each image set. 
As the number of randomly selected images increases, more and 
more estimated relative rotations will become
outliers, and thus it poses a greater challenge for absolute rotation
estimation. The results are reported in \Cref{fig:conf_noise}.
As seen from this figure, with the increasing amounts of outliers, the performance
of the `w/o confidence' variant becomes much poorer, while the full EAR-Net
is not sensitive to outliers mainly because low confidences are 
automatically assigned to the outliers. 

Thirdly, we analyze the relationship between the errors of the predicted
relative rotations and their confidences. We firstly sample around 230k
image pairs from the ScanNet dataset, and
compute the relative rotations and their confidences accordingly.
Then, we divide the confidences in [0, 1] into 20 groups with
equal intervals, and for each group, the mean and median errors of the
predicted relative rotations are computed using the ground-truth relative rotations. 
The mean and median errors of the relative rotations in these groups are plotted
in \Cref{fig:err_conf}. As seen from this figure, the relative
rotation error tends to drop when the confidence score increases.
The error is significantly larger when the confidence score is
close to zero, which is possibly because lots of
outliers occur near this area. 

\begin{figure*}[t]
  \centering
  \includegraphics[width=0.6\linewidth,clip]{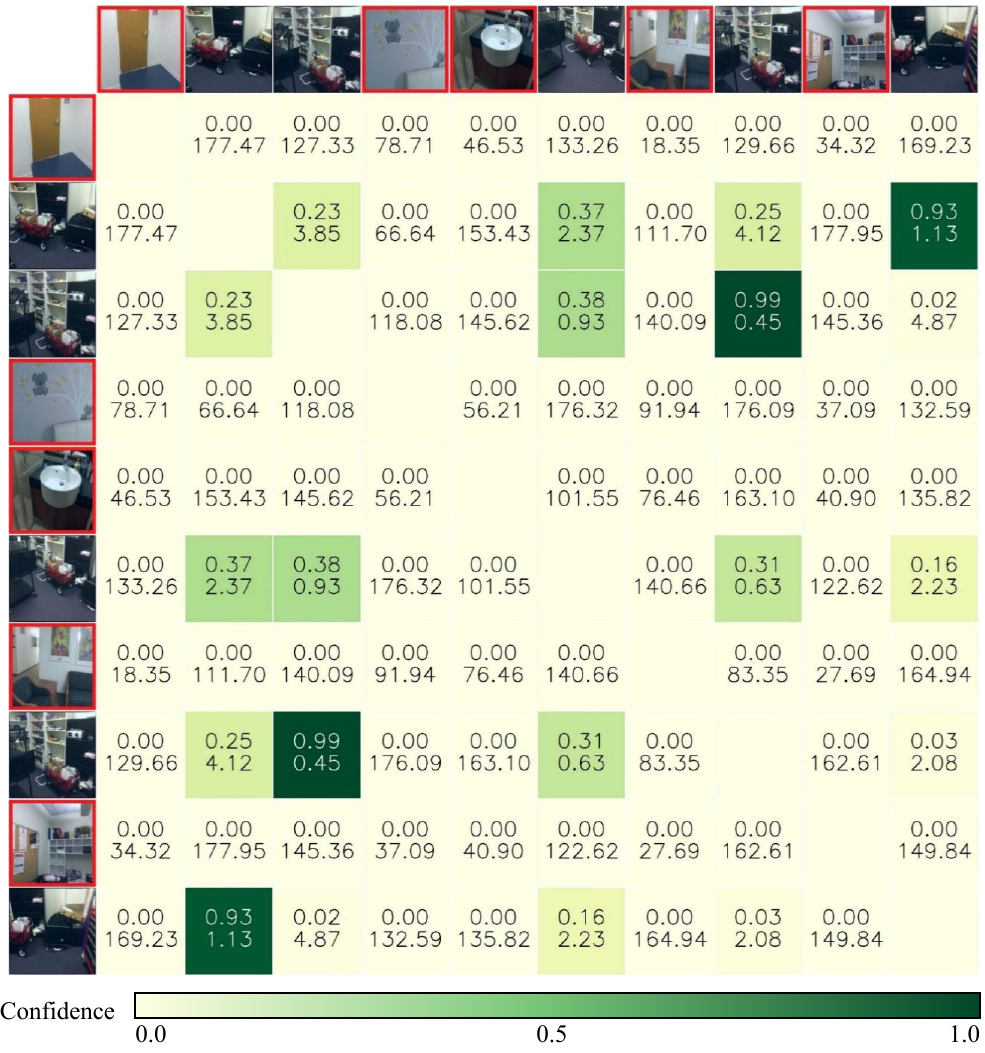}
  \caption{Visualization of the pairwise relative rotation errors 
    and their corresponding confidence scores predicted by the dual-branch decoder 
    on corrupted data from ScanNet.
    The numbers denote the confidence (top) and the error (bottom, in degree).
    The images marked with \textcolor[RGB]{236,43,39}{red} 
    border denote the noise images that have no overlap with others.
    Our model predicts close-to-zero confidences for those relative rotations 
    to the outliers, which could alleviate their negative influence.
  }
  \label{fig:weight_viz_noise}
\end{figure*}

Moreover, in \Cref{fig:weight_viz}, we visualize
a set of images and the errors of the predicted pairwise relative
rotations as well as the corresponding confidences.
As seen from this figure, EAR-Net tends to predict close-to-zero
confidences for image pairs with large rotation errors, 
\eg, the 1st-6th image pair (the confidence 
is around 0 and the error is 20.3\degree), and large confidences 
for image pairs with small errors, \eg, the 3rd-5th image 
(the confidence is 0.81 and the error is 1.14\degree). 
Moreover, image pairs with small/large overlap areas generally have
small/large confidences, possibly because the model could give
a more reliable estimation when image pairs have a larger overlap area.
For example, the 3rd-5th image pair has a large overlap area, and 
the corresponding confidence is around 0.81. The 5th-6th image pair 
has little overlap, and the corresponding confidence is close to zero. 
The above results confirm that reliable/unreliable 
relative rotations tend to be assigned with large/small confidences, 
as indicated in \Cref{subsec:graph}. 
In addition, we also visualize the case when image sets are corrupted. Specifically,
five images are normally sampled and another five images are
randomly selected (noise images) from other scenes with no overlap.
\cref{fig:weight_viz_noise} shows the corresponding results.
As seen from this figure, the confidence scores between the noise images
and other images are close to zero, which could alleviate
the negative influence of outlier edges via the proposed
confidence-aware initialization approach
and confidence-aware optimization algorithm.

\noindent\textbf{Effect of CAI. }
In order to investigate the effect of the CAI (confidence-aware initialization) approach, 
we evaluate the proposed EAR-Net by replacing the CAI approach with random 
initialization (w/o CAI). 
We test the `w/o CAI' variant on the ScanNet dataset five times independently, 
and the corresponding average results are reported in \cref{tab:ablation_scannet}.
As seen from this table,  
the `w/o CAI' variant performs worse than the full EAR-Net, 
demonstrating the effectiveness of the designed CAI approach.

In addition, we also investigate the effect of the CAI approach 
for resisting outliers by corrupting the image sets by appending another
$\{0,2,4,6,8,10\}$ randomly
selected images from other scenes with no overlap to each image set. 
This setup is the same as evaluating the effect of the confidence for resisting outliers.
The results are shown in \Cref{fig:CAI}.
As seen from this figure, as the outlier level increases, 
EAR-Net maintains a stable performance, 
while the performance of the `w/o CAI' variant is
degraded severely. This indicates that the proposed CAI approach is
essential for absolute rotation estimation.

\begin{table}[tb]
  \centering
  \fs
  \def\len{0.6cm}
  \def\degree{$^\circ$}
  \begin{tabular}{L{2.0cm}C{2.07cm}C{\len}C{\len}C{1.35cm}}
    \toprule
    Method             & Loss Formula     & Mn$\downarrow$ & Med$\downarrow$ & Acc@10\degree$\uparrow$\\
    \midrule
    EAR-Net-$\ell_2$   &  $\frac{x^2}{2}$                                      & 9.31 & 3.85 & 78.71 \\
    EAR-Net-Cau        &  $\frac{\alpha^2}{2}\log(1+\frac{x^2}{\alpha^2})$     & 7.18 & 3.14 & 85.39 \\
    EAR-Net-GM         &  $\frac{x^2}{2(\alpha^2+x^2)}$                        & 7.36 & 3.19 & 84.82 \\
    EAR-Net-CAL        &  $cx^2$                                     & \textbf{4.03} & \textbf{2.06} & \textbf{94.18} \\
    \bottomrule
  \end{tabular}
  \caption{Comparison of different loss functions on the 
  ScanNet dataset \cite{ScanNet}. `GM' denotes the Geman-McClure loss.
  `Cau' denotes the Cauchy loss.
    `Mn' and `Med' denote the mean and median error respectively.
  }
  \label{tab:robust_cost}
\end{table}

\noindent\textbf{Comparison of Different Loss Functions. } 
As indicated in \cite{IRLS}, robust loss functions are helpful for dealing with 
outliers. Hence, we further evaluate the proposed EAR-Net by
replacing the proposed CAL in \cref{eqn:objective} with the following loss
functions respectively:
(1) The naive $\ell_2$ loss, which serves as a baseline (EAR-Net-$\ell_2$).
(2) The Cauchy loss function (EAR-Net-Cau).
(3) The Geman-McClure loss function (EAR-Net-GM).
The formulations of the above functions are summarized in the second column of \cref{tab:robust_cost}.
We set $\alpha=5^\circ$ in the Cauchy and Geman-McClure loss functions as 
suggested by Chatterjee \etal{} \cite{IRLS}.

The corresponding results are reported in \Cref{tab:robust_cost}.
As seen from this table, incorporating the Cauchy and Geman-McClure
loss functions leads to improved model performance compared with using
the naive $\ell_2$ loss, consistent with the observations in \cite{IRLS}.
In addition, it is noted that the
EAR-Net with the proposed CAL achieves a
significantly higher accuracy than all the evaluated
model variants with different robust loss functions, demonstrating 
directly learning the weights from data is more beneficial.

\begin{table}
  \centering
  \fs
  \def\len{0.8cm}
  \begin{tabular}{lccc}
    \toprule
    Method & Mn$\downarrow$ & Med$\downarrow$ & Acc@10\degree$\uparrow$\\
    \midrule
    Reg6D+RAGO                        & 11.87 & 6.55 & 67.37 \\
    Reg6D$\rightarrow$RAGO            & 11.57 & 5.72 & 71.53 \\
    \midrule  
    ExtremeRotation+RAGO              & 9.83 & 5.04 & 75.26 \\
    ExtremeRotation$\rightarrow$RAGO  & 9.04 & 4.82 & 76.77 \\
    \midrule
    EAR-Net                           & \textbf{4.03} & \textbf{2.06} & \textbf{94.18} \\
    \bottomrule
  \end{tabular}
  \caption{Comparison of different estimation strategies for predicting absolute rotations on ScanNet.
          `Mn' and `Med' denote the mean and median error respectively.
          `A+B' denotes A and B are trained respectively in a two-stage manner.
          `A$\rightarrow$B'  denotes A and B are combined and jointly trained 
          together in an end-to-end manner.
  }
  \label{tab:concat}
\end{table}

\subsection{EAR-Net vs End-to-End Learning by Combining Existing Techniques}
\label{subsec:concat}

It is noted that when a learning-based technique for predicting absolute rotations 
from relative rotations (\eg{}, RAGO \cite{RAGO}) is combined and jointly trained 
with a learning-based relative rotation estimation technique 
(\eg{}, Reg6D \cite{rot6d}, ExtremeRotation \cite{Cai2021extreme})  
from input images, we could straightforwardly obtain an end-to-end method for 
predicting absolute rotations from input images. 
Accordingly in this subsection, we train Reg6D$\rightarrow$RAGO (combining Reg6D and RAGO) 
and ExtremeRotation$\rightarrow$RAGO (combining ExtremeRotation and RAGO) 
respectively in an end-to-end training manner 
where only the ground-truth absolute rotations in the training set are used as supervision 
signals as done in the proposed method, and the corresponding results on the 
ScanNet dataset \cite{ScanNet} are reported in \cref{tab:concat}.
For a clear comparison, \cref{tab:concat} also reports the results of the 
proposed method as well as the results 
(that are cited from \cref{tab:scannet}) by training  Reg6D and RAGO 
(also ExtremeRotation and RAGO)  in a two-stage manner, \ie{}, 
the relative rotation estimation method 
Reg6D (or ExtremeRotation) is trained firstly by utilizing ground 
truth relative rotations as supervision signals, 
and then RAGO is trained by utilizing ground-truth absolute rotations as supervision signals.

Two points could be observed from this table: 
(i) Both Reg6D$\rightarrow$RAGO and ExtremeRotation$\rightarrow$RAGO 
perform better than their two-stage counterparts, demonstrating that end-to-end training 
is also effective for boosting the performance of existing models.
(ii) The two end-to-end methods Reg6D$\rightarrow$RAGO and ExtremeRotation$\rightarrow$RAGO perform 
significantly worse than the proposed method EAR-Net, 
demonstrating that such a simple combination of existing 
techniques could not guarantee a competitive performance. 

\subsection{Integration of Existing Confidence Estimation Methods}
\label{subsec:relconf}

In this subsection, we investigate the integration of existing confidence estimation methods
into the proposed EAR-Net.
Specifically, it is noted that several existing works for 3D reconstruction \cite{wang2024dust3r,leroy2024mast3r,wang2025continuous} 
also estimate confidence, and
they are trained with the following confidence regression objective function:
\begin{equation}
\mathcal{L}_{conf}=c\|x-\hat{x}\|_2-\alpha\log{c},
\label{eqn:conf}
\end{equation}
where $x$ denotes the predicted 3D point, $\hat{x}$ denotes the ground-truth 3D point, 
$c$ denotes the predicted confidence of $x$, and $\alpha$ is a preset parameter.
As seen from this objective, a larger point confidence indicates that the 3D position 
of a reconstructed point is closer to its ground-truth position.

\begin{table}[t]
\fs
\def\degree{$^\circ$}
    \centering
    \begin{tabular}{lcccc}
    \toprule
    Method & Mn$\downarrow$ & Med$\downarrow$     & Acc@10\degree$\uparrow$ \\
    \midrule
    EAR-Net$^\dagger$      & 7.07 & 3.17 & 83.02  \\
    EAR-Net          & \bf 4.03 & \bf 2.06 & \bf 94.18  \\
    \bottomrule
    \end{tabular}
    \vspace{-0,1in}
    \caption{Comparison between the integrated version (EAR-Net$^\dagger$) 
    and the original version (EAR-Net) for absolute rotation estimation on ScanNet.}
    \label{tab:conf}
\end{table}

\begin{table*}[t]
\fs
\def\degree{$^\circ$}
\centering
\begin{tabular}{lcccc}
\toprule
\multirow{1}{*}{Method} & Mn$\downarrow$ & Med$\downarrow$ & Acc@10\degree$\uparrow$ & NS/NT$\uparrow$ \\
\midrule
SIFT+HARA            & 65.53 & 49.52 & 10.64 & 1132/1485 \\
SuperPoint+HARA      & 54.50 & 36.87 & 13.62 & 1288/1485 \\
LoFTR+HARA           & 39.09 & 25.95 & 17.64 & 1079/1485 \\
ASpanFormer+HARA     & 38.24 & 26.80 & 17.05 & 1483/1485 \\
Reg6D+HARA           & 36.52 & 23.97 & 22.87 & \textbf{1485}/1485 \\
ExtremeRotation+HARA & 37.23 & 23.40 & 22.17 & \textbf{1485}/1485 \\
\midrule
PoseDiffusion        & 40.94 & 28.70 & 21.21 & \textbf{1485}/1485 \\
RelPose++            & 36.45 & 21.28 & 30.84 & \textbf{1485}/1485 \\
PRAGO                & 34.41 & 20.94 & 24.05 & \textbf{1485}/1485 \\
\midrule
EAR-Net              & \textbf{21.97} & \textbf{9.46} & \textbf{52.04} & \textbf{1485}/1485 \\
\bottomrule
\end{tabular}
\vspace{-0.1in}
\caption{Performance comparison under the low-overlap setup with overlap ratios of [0.1, 0.2] on ScanNet.}
\label{tab:lowoverlap}
\end{table*}

Here, Function (\ref{eqn:conf}) is integrated with the proposed method for 
learning relative-rotation confidence, and its variables \{$x$, $\hat{x}$\} represent 
the predicted and ground-truth relative rotation, rather than the predicted and 
ground-truth 3D point as done in \cite{wang2024dust3r,leroy2024mast3r,wang2025continuous}. 
Accordingly, the relative rotations and their confidences could also be jointly learned by optimizing 
Function (\ref{eqn:conf}). 
Once the relative rotations and their confidence scores are obtained, they could be 
straightforwardly inputted into our downstream confidence-aware rotation averaging module 
for absolute rotation estimation. We denote this integrated version as EAR-Net$^\dagger$.

As done in \cref{subsec:evaluation}, we trained and evaluated 
EAR-Net$^\dagger$ on ScanNet \cite{ScanNet}, 
and the corresponding results are reported in \cref{tab:conf}. 
As seen from \cref{tab:conf,tab:scannet}, the integrated version EAR-Net$^\dagger$ 
performs better than most of the comparative existing methods, 
demonstrating that it is feasible to integrate the objective function (\ref{eqn:conf}) with our pipeline. 
Moreover, the original EAR-Net performs much better than the integrated version, 
demonstrating the effectiveness of our designed confidence estimation manner. 

\subsection{Limitation}
\label{subsec:limit}

In this subsection, we discuss the limitation of our method. 
In theory, since the proposed method extracts the feature maps from input images to learn pairwise 
features for constructing the introduced epipolar confidence graph as illustrated in 
\cref{fig:architecture}, the overlap ratio among these images is an important 
factor for influencing the performance of the proposed method: the performance of the proposed method 
would degrade when the overlap ratio among the input images is lowered down. 
This is a theoretical limitation of the proposed method, while in fact, most of the comparative methods 
(e.g., \cite{SIFT,SuperPoint}) have to suffer from the same limitation as 
indicated in \cite{Cai2021extreme}.

To quantitatively analyze the above theoretical limitation, we have conducted the following experiment on ScanNet. Specifically, we sampled image sets with overlap ratios of [0.1, 0.2] 
(much lower than the original overlap ratios of [0.4, 0.8] in \cref{subsec:evaluation}), 
resulting in 1485 image sets in total. 
We then evaluate the proposed EAR-Net as well as 9 existing comparative methods 
(that perform relatively better than the others).
The corresponding results are reported in \cref{tab:lowoverlap}. 
Two points could be observed from this table and \cref{tab:scannet}:
(i)  When the overlap ratio is lowered down from [0.4, 0.8] to [0.1, 0.2], 
the performance of all the involved methods degrades significantly, 
consistent with the aforementioned theoretical limitation. 
(ii) Under the low overlap setup, although the performance of the proposed method 
degrades significantly as indicated in Point i, 
it still shows remarkable superiority to the other comparative methods. 

Considering that the overlap ratio among the input images 
has an important influence on the performance of both the proposed method 
and other comparative methods as discussed above,
a future direction for improvement is to investigate how to learn 
more effective and robust features from input images for rotation estimation in low-overlap cases.

\section{Conclusion}
\label{sec:conclusion}

Unlike existing methods that adopt a multi-stage strategy 
which inevitably leads to the accumulation of error caused by each 
involved operation, this paper proposes the end-to-end EAR-Net 
for recovering absolute rotations from multi-view images directly. 
The EAR-Net consists of two key modules, including the 
epipolar-confidence graph construction module and 
the confidence-aware rotation averaging module. 
The epipolar-confidence graph construction module is explored 
to predict relative rotations among the input images 
and their confidences, which results in the epipolar confidence graph. 
Then the confidence-aware rotation averaging module takes this 
graph as input, and it outputs the estimated absolute rotations by minimizing the 
proposed confidence-aware loss via the proposed confidence-aware initialization 
approach and the confidence-aware optimization algorithm. 
Extensive experimental results on three public datasets demonstrate the 
effectiveness of the proposed EAR-Net in terms of both accuracy and inference speed.

\bibliographystyle{IEEEtran}
\bibliography{ref}

@inproceedings{detr,
  title={End-to-end object detection with transformers},
  author={Carion, Nicolas and Massa, Francisco and Synnaeve, Gabriel and Usunier, Nicolas and Kirillov, Alexander and Zagoruyko, Sergey},
  booktitle={European Conference on Computer Vision},
  pages={213--229},
  year={2020},
  organization={Springer}
}

@inproceedings{epro,
  title={EPro-PnP: Generalized End-to-End Probabilistic Perspective-n-Points for Monocular Object Pose Estimation},
  author={Chen, Hansheng and Wang, Pichao and Wang, Fan and Tian, Wei and Xiong, Lu and Li, Hao},
  booktitle={Proceedings of the IEEE Conference on Computer Vision and Pattern Recognition},
  pages={2781--2790},
  year={2022}
}

@inproceedings{ResNet,
  title={Identity mappings in deep residual networks},
  author={He, Kaiming and Zhang, Xiangyu and Ren, Shaoqing and Sun, Jian},
  booktitle={European Conference on Computer Vision},
  pages={630--645},
  year={2016},
  organization={Springer}
}

@inproceedings{colmap,
  title={Structure-from-motion revisited},
  author={Schonberger, Johannes L and Frahm, Jan-Michael},
  booktitle={Proceedings of the IEEE Conference on Computer Vision and Pattern Recognition},
  pages={4104--4113},
  year={2016}
}

@article{dong2022,
    author = {Dong, Qiulei and Gao, Xiang and Cui, Hainan and hu, Zhanyi},
    year = {2022},
    month = {05},
    pages = {862-872},
    title = {Robust Camera Translation Estimation via Rank Enforcement},
    volume = {52},
    number={2},
    journal = {IEEE Transactions on Cybernetics},
    doi = {10.1109/TCYB.2020.2988679}
}

@incollection{snavely2006photo,
  title={Photo tourism: exploring photo collections in 3D},
  author={Snavely, Noah and Seitz, Steven M and Szeliski, Richard},
  booktitle={ACM Siggraph},
  pages={835--846},
  year={2006}
}

@inproceedings{mvsnet,
  title={Mvsnet: Depth inference for unstructured multi-view stereo},
  author={Yao, Yao and Luo, Zixin and Li, Shiwei and Fang, Tian and Quan, Long},
  booktitle={European Conference on Computer Vision},
  pages={767--783},
  year={2018}
}

@inproceedings{bae2022multi,
  title={Multi-View Depth Estimation by Fusing Single-View Depth Probability with Multi-View Geometry},
  author={Bae, Gwangbin and Budvytis, Ignas and Cipolla, Roberto},
  booktitle={Proceedings of the IEEE Conference on Computer Vision and Pattern Recognition},
  pages={2842--2851},
  year={2022}
}

@inproceedings{melekhov2017relative,
  title={Relative camera pose estimation using convolutional neural networks},
  author={Melekhov, Iaroslav and Ylioinas, Juha and Kannala, Juho and Rahtu, Esa},
  booktitle={International Conference on Advanced Concepts for Intelligent Vision Systems},
  pages={675--687},
  year={2017},
  organization={Springer}
}

@inproceedings{en2018rpnet,
  title={Rpnet: An end-to-end network for relative camera pose estimation},
  author={En, Sovann and Lechervy, Alexis and Jurie, Fr{\'e}d{\'e}ric},
  booktitle={Proceedings of the European Conference on Computer Vision Workshops},
  year={2018}
}

@InProceedings{Cai2021extreme,
  author    = {Cai, Ruojin and Hariharan, Bharath and Snavely, Noah and Averbuch-Elor, Hadar},
  title     = {Extreme Rotation Estimation Using Dense Correlation Volumes},
  booktitle = {Proceedings of the IEEE Conference on Computer Vision and Pattern Recognition},
  year      = {2021},
  pages     = {14566-14575}
}

@inproceedings{rot6d,
  title={On the continuity of rotation representations in neural networks},
  author={Zhou, Yi and Barnes, Connelly and Lu, Jingwan and Yang, Jimei and Li, Hao},
  booktitle={Proceedings of the IEEE Conference on Computer Vision and Pattern Recognition},
  pages={5745--5753},
  year={2019}
}

@article{nister2004efficient,
title={An efficient solution to the five-point relative pose problem},
author={Nist{\'e}r, David},
journal={IEEE Transactions on Pattern Analysis and Machine Intelligence},
volume={26},
number={6},
pages={756--770},
year={2004},
publisher={IEEE}
}

@InProceedings{RANSACRole,
  author    = {Fan, Hongyi and Kileel, Joe and Kimia, Benjamin},
  title     = {On the Instability of Relative Pose Estimation and RANSAC's Role},
  booktitle = {Proceedings of the IEEE Conference on Computer Vision and Pattern Recognition},
  month     = {June},
  year      = {2022},
  pages     = {8935-8943}
}

@inproceedings{allinweights,
  title={It is all in the weights: robust rotation averaging revisited},
  author={Sidhartha, Chitturi and Govindu, Venu Madhav},
  booktitle={International Conference on 3D Vision},
  pages={1134--1143},
  year={2021},
}

@InProceedings{HARA,
    author    = {Lee, Seong Hun and Civera, Javier},
    title     = {HARA: A Hierarchical Approach for Robust Rotation Averaging},
    booktitle = {Proceedings of the IEEE Conference on Computer Vision and Pattern Recognition},
    month     = {June},
    year      = {2022},
    pages     = {15777-15786}
}

@inproceedings{NeuRoRA,
  title={Neurora: Neural robust rotation averaging},
  author={Purkait, Pulak and Chin, Tat-Jun and Reid, Ian},
  booktitle={European Conference on Computer Vision},
  pages={137--154},
  year={2020},
}

@InProceedings{RAGO,
    author    = {Li, Heng and Cui, Zhaopeng and Liu, Shuaicheng and Tan, Ping},
    title     = {RAGO: Recurrent Graph Optimizer for Multiple Rotation Averaging},
    booktitle = {Proceedings of the IEEE Conference on Computer Vision and Pattern Recognition},
    month     = {June},
    year      = {2022},
    pages     = {15787-15796}
}

@InProceedings{Chen_2021_CVPR,
    author    = {Chen, Yu and Zhao, Ji and Kneip, Laurent},
    title     = {Hybrid Rotation Averaging: A Fast and Robust Rotation Averaging Approach},
    booktitle = {Proceedings of the IEEE Conference on Computer Vision and Pattern Recognition},
    month     = {June},
    year      = {2021},
    pages     = {10358-10367}
}

@InProceedings{RAMSP,
    author    = {Yang, Luwei and Li, Heng and Rahim, Jamal Ahmed and Cui, Zhaopeng and Tan, Ping},
    title     = {End-to-End Rotation Averaging With Multi-Source Propagation},
    booktitle = {Proceedings of the IEEE Conference on Computer Vision and Pattern Recognition},
    month     = {June},
    year      = {2021},
    pages     = {11774-11783}
}

@article{IRLS,
  title={Robust relative rotation averaging},
  author={Chatterjee, Avishek and Govindu, Venu Madhav},
  journal={IEEE Transactions on Pattern Analysis and Machine Intelligence},
  volume={40},
  number={4},
  pages={958--972},
  year={2017},
}

@inproceedings{hartleyAvg,
  title={L1 rotation averaging using the Weiszfeld algorithm},
  author={Hartley, Richard and Aftab, Khurrum and Trumpf, Jochen},
  booktitle={Proceedings of the IEEE Conference on Computer Vision and Pattern Recognition},
  pages={3041--3048},
  year={2011},
}

@inproceedings{DISCO,
  title={Discrete-continuous optimization for large-scale structure from motion},
  author={Crandall, David and Owens, Andrew and Snavely, Noah and Huttenlocher, Dan},
  booktitle={Proceedings of the IEEE Conference on Computer Vision and Pattern Recognition},
  pages={3001--3008},
  year={2011},
  organization={IEEE}
}

@inproceedings{govindu2006robustness,
  title={Robustness in motion averaging},
  author={Govindu, Venu Madhav},
  booktitle={Asian Conference on Computer Vision},
  pages={457--466},
  year={2006},
  organization={Springer}
}

@inproceedings{govindu2004lie,
  title={Lie-algebraic averaging for globally consistent motion estimation},
  author={Govindu, Venu Madhav},
  booktitle={Proceedings of the IEEE Computer Society Conference on Computer Vision and Pattern Recognition},
  volume={1},
  year={2004},
  organization={IEEE}
}

@inproceedings{zach2010disambiguating,
  title={Disambiguating visual relations using loop constraints},
  author={Zach, Christopher and Klopschitz, Manfred and Pollefeys, Marc},
  booktitle={Proceedings of the IEEE Conference on Computer Vision and Pattern Recognition},
  pages={1426--1433},
  year={2010},
  organization={IEEE}
}

@inproceedings{cui2018voting,
  title={Voting-based incremental structure-from-motion},
  author={Cui, Hainan and Shen, Shuhan and Gao, Wei},
  booktitle={Proceedings of the IEEE International Conference on Pattern Recognition},
  pages={1929--1934},
  year={2018},
}

@inproceedings{shen2016graph,
  title={Graph-based consistent matching for structure-from-motion},
  author={Shen, Tianwei and Zhu, Siyu and Fang, Tian and Zhang, Runze and Quan, Long},
  booktitle={European Conference on Computer Vision},
  pages={139--155},
  year={2016},
  organization={Springer}
}

@article{gao2021incremental,
  title={Incremental rotation averaging},
  author={Gao, Xiang and Zhu, Lingjie and Xie, Zexiao and Liu, Hongmin and Shen, Shuhan},
  journal={International Journal of Computer Vision},
  volume={129},
  number={4},
  pages={1202--1216},
  year={2021},
  publisher={Springer}
}

@article{chen2020graph,
  title={Graph-based parallel large scale structure from motion},
  author={Chen, Yu and Shen, Shuhan and Chen, Yisong and Wang, Guoping},
  journal={Pattern Recognition},
  volume={107},
  pages={107537},
  year={2020},
  publisher={Elsevier}
}

@inproceedings{keynet,
  title={Key. net: Keypoint detection by handcrafted and learned cnn filters},
  author={Barroso-Laguna, Axel and Riba, Edgar and Ponsa, Daniel and Mikolajczyk, Krystian},
  booktitle={Proceedings of the IEEE International Conference on Computer Vision},
  pages={5836--5844},
  year={2019}
}

@InProceedings{SuperGlue,
  author = {Sarlin, Paul-Edouard and DeTone, Daniel and Malisiewicz, Tomasz and Rabinovich, Andrew},
  title = {SuperGlue: Learning Feature Matching With Graph Neural Networks},
  booktitle = {IEEE Conference on Computer Vision and Pattern Recognition},
  month = {June},
  year = {2020}
}

@InProceedings{LoFTR,
    author    = {Sun, Jiaming and Shen, Zehong and Wang, Yuang and Bao, Hujun and Zhou, Xiaowei},
    title     = {LoFTR: Detector-Free Local Feature Matching With Transformers},
    booktitle = {Proceedings of the IEEE Conference on Computer Vision and Pattern Recognition},
    month     = {June},
    year      = {2021},
    pages     = {8922-8931}
}

@inproceedings{aspanformer,
  title={Aspanformer: Detector-free image matching with adaptive span transformer},
  author={Chen, Hongkai and Luo, Zixin and Zhou, Lei and Tian, Yurun and Zhen, Mingmin and Fang, Tian and Mckinnon, David and Tsin, Yanghai and Quan, Long},
  booktitle={European Conference on Computer Vision},
  pages={20--36},
  year={2022},
  organization={Springer}
}

@InProceedings{patch2pix,
    author    = {Zhou, Qunjie and Sattler, Torsten and Leal-Taixe, Laura},
    title     = {Patch2Pix: Epipolar-Guided Pixel-Level Correspondences},
    booktitle = {Proceedings of the IEEE Conference on Computer Vision and Pattern Recognition},
    month     = {June},
    year      = {2021},
    pages     = {4669-4678}
}

@inproceedings{matchformer,
  title={MatchFormer: Interleaving Attention in Transformers for Feature Matching},
  author={Wang, Qing and Zhang, Jiaming and Yang, Kailun and Peng, Kunyu and Stiefelhagen, Rainer},
  booktitle={Asian Conference on Computer Vision},
  year={2022}
}

@article{R2D2,
  title={R2d2: Reliable and repeatable detector and descriptor},
  author={Revaud, Jerome and De Souza, Cesar and Humenberger, Martin and Weinzaepfel, Philippe},
  journal={Advances in Neural Information Processing Systems},
  volume={32},
  year={2019}
}

@ARTICLE{SIFT,
  author = {David G. Lowe},
  title = {Distinctive Image Features from Scale-Invariant Keypoints},
  journal = {International Journal of Computer Vision},
  volume={60},
  number={2},
  pages={91--110},
  year={2004},
  publisher={Springer}
}

@inproceedings{SuperPoint,
  title={Superpoint: Self-supervised interest point detection and description},
  author={DeTone, Daniel and Malisiewicz, Tomasz and Rabinovich, Andrew},
  booktitle={Proceedings of the IEEE Conference on Computer Vision and Pattern Recognition workshops},
  pages={224--236},
  year={2018}
}

@inproceedings{HyNet,
  Author = {Yurun Tian and Axel Barroso-Laguna and Tony Ng and Vassileios Balntas and Krystian Mikolajczyk},
  Title = {HyNet: Learning Local Descriptor with Hybrid Similarity Measure and Triplet Loss},
  Year = {2020},
  volume={33},
  pages={7401--7412},
  booktitle = {Advances in Neural Information Processing Systems},
}

@inproceedings{ScanNet,
    title={ScanNet: Richly-annotated 3D Reconstructions of Indoor Scenes},
    author={Dai, Angela and Chang, Angel X. and Savva, Manolis and Halber, Maciej and Funkhouser, Thomas and Nie{\ss}ner, Matthias},
    booktitle = {Proceedings of the IEEE Conference on Computer Vision and Pattern Recognition},
    year = {2017}
}

@article{DTU,
  title={Large-scale data for multiple-view stereopsis},
  author={Aan{\ae}s, Henrik and Jensen, Rasmus Ramsb{\o}l and Vogiatzis, George and Tola, Engin and Dahl, Anders Bjorholm},
  journal={International Journal of Computer Vision},
  volume={120},
  number={2},
  pages={153--168},
  year={2016},
  publisher={Springer}
}

@INPROCEEDINGS{7Scene,
  author={Shotton, Jamie and Glocker, Ben and Zach, Christopher and Izadi, Shahram and Criminisi, Antonio and Fitzgibbon, Andrew},
  booktitle={Proceedings of the IEEE Conference on Computer Vision and Pattern Recognition}, 
  title={Scene Coordinate Regression Forests for Camera Relocalization in RGB-D Images}, 
  year={2013},
  pages={2930-2937},
  doi={10.1109/CVPR.2013.377}
  }

@article{RANSAC,
  title={Random sample consensus: a paradigm for model fitting with applications to image analysis and automated cartography},
  author={Fischler, Martin A and Bolles, Robert C},
  journal={Communications of the ACM},
  volume={24},
  number={6},
  pages={381--395},
  year={1981},
  publisher={ACM New York, NY, USA}
}

@article{quaternionAvg,
  title={Averaging quaternions},
  author={Markley, F Landis and Cheng, Yang and Crassidis, John L and Oshman, Yaakov},
  journal={Journal of Guidance, Control, and Dynamics},
  volume={30},
  number={4},
  pages={1193--1197},
  year={2007}
}

@inproceedings{coco,
  title={Microsoft coco: Common objects in context},
  author={Lin, Tsung-Yi and Maire, Michael and Belongie, Serge and Hays, James and Perona, Pietro and Ramanan, Deva and Doll{\'a}r, Piotr and Zitnick, C Lawrence},
  booktitle={European Conference on Computer Vision},
  pages={740--755},
  year={2014},
  organization={Springer}
}

@article{mildenhall2021nerf,
  title={Nerf: Representing scenes as neural radiance fields for view synthesis},
  author={Mildenhall, Ben and Srinivasan, Pratul P and Tancik, Matthew and Barron, Jonathan T and Ramamoorthi, Ravi and Ng, Ren},
  journal={Communications of the ACM},
  volume={65},
  number={1},
  pages={99--106},
  year={2021},
  publisher={ACM New York, NY, USA}
}

@inproceedings{chen2021wide,
  title={Wide-baseline relative camera pose estimation with directional learning},
  author={Chen, Kefan and Snavely, Noah and Makadia, Ameesh},
  booktitle={Proceedings of the IEEE Conference on Computer Vision and Pattern Recognition},
  pages={3258--3268},
  year={2021}
}

@article{dsac,
  title={Visual camera re-localization from RGB and RGB-D images using DSAC},
  author={Brachmann, Eric and Rother, Carsten},
  journal={IEEE Transactions on Pattern Analysis and Machine Intelligence},
  volume={44},
  number={9},
  pages={5847--5865},
  year={2021},
  publisher={IEEE}
}

@inproceedings{yi2016lift,
  title={Lift: Learned invariant feature transform},
  author={Yi, Kwang Moo and Trulls, Eduard and Lepetit, Vincent and Fua, Pascal},
  booktitle={European Conference on Computer Vision},
  pages={467--483},
  year={2016},
  organization={Springer}
}

@inproceedings{wang2023posediffusion,
  title={Posediffusion: Solving pose estimation via diffusion-aided bundle adjustment},
  author={Wang, Jianyuan and Rupprecht, Christian and Novotny, David},
  booktitle={Proceedings of the IEEE International Conference on Computer Vision},
  pages={9773--9783},
  year={2023}
}

@inproceedings{taiana2024prago,
  title={PRAGO: Differentiable Multi-View Pose Optimization From Objectness Detections},
  author={Taiana, Matteo and Toso, Matteo and James, Stuart and Del Bue, Alessio},
  booktitle={International Conference on 3D Vision},
  pages={324--333},
  year={2024},
}

@inproceedings{lin2024relpose++,
  title={Relpose++: Recovering 6d poses from sparse-view observations},
  author={Lin, Amy and Zhang, Jason Y and Ramanan, Deva and Tulsiani, Shubham},
  booktitle={International Conference on 3D Vision},
  pages={106--115},
  year={2024},
}

@inproceedings{wang2024dust3r,
  title={Dust3r: Geometric 3d vision made easy},
  author={Wang, Shuzhe and Leroy, Vincent and Cabon, Yohann and Chidlovskii, Boris and Revaud, Jerome},
  booktitle={Proceedings of the IEEE Conference on Computer Vision and Pattern Recognition},
  pages={20697--20709},
  year={2024}
}

@inproceedings{leroy2024mast3r,
  title={Grounding image matching in 3d with mast3r},
  author={Leroy, Vincent and Cabon, Yohann and Revaud, J{\'e}r{\^o}me},
  booktitle={European Conference on Computer Vision},
  pages={71--91},
  year={2024},
  organization={Springer}
}

@inproceedings{wang2025continuous,
  title={Continuous 3d perception model with persistent state},
  author={Wang, Qianqian and Zhang, Yifei and Holynski, Aleksander and Efros, Alexei A and Kanazawa, Angjoo},
  booktitle={Proceedings of the IEEE Conference on Computer Vision and Pattern Recognition},
  pages={10510--10522},
  year={2025}
}

@inproceedings{wang2025vggt,
  title={Vggt: Visual geometry grounded transformer},
  author={Wang, Jianyuan and Chen, Minghao and Karaev, Nikita and Vedaldi, Andrea and Rupprecht, Christian and Novotny, David},
  booktitle={Proceedings of the Computer Vision and Pattern Recognition Conference},
  pages={5294--5306},
  year={2025}
}

@article{kim2021oln,
  title={Learning Open-World Object Proposals without Learning to Classify},
  author={Kim, Dahun and Lin, Tsung-Yi and Angelova, Anelia and Kweon, In So and Kuo, Weicheng},
  journal={IEEE Robotics and Automation Letters (RA-L)},
  year={2022}
}

@inproceedings{yeshwanth2023scannet++,
  title={Scannet++: A high-fidelity dataset of 3d indoor scenes},
  author={Yeshwanth, Chandan and Liu, Yueh-Cheng and Nie{\ss}ner, Matthias and Dai, Angela},
  booktitle={Proceedings of the IEEE International Conference on Computer Vision},
  pages={12--22},
  year={2023}
}

\end{document}